# A General Albedo Recovery Approach for Aerial Photogrammetric Images through Inverse Rendering

Shuang Song [1,2,4], Rongjun Qin [1,2,3,4] *

[1] Geospatial Data Analytics Laboratory, The Ohio State University, Columbus, USA

[2] Department of Civil, Environmental and Geodetic Engineering, The Ohio State University, Columbus, USA

[3] Department of Electrical and Computer Engineering, The Ohio State University, Columbus, USA

[4] Translational Data Analytics Institute, The Ohio State University, Columbus, USA

*Corresponding author: qin.324@osu.edu

**Abstract:** Modeling outdoor scenes for the synthetic 3D environment requires the recovery of reflectance/albedo information from raw images, which is an ill-posed problem due to the complicated unmodeled physics in this process (e.g., indirect lighting, volume scattering, specular reflection). The problem remains unsolved in a practical context. The recovered albedo can facilitate model relighting and shading, which can further enhance the realism of rendered models and the applications of digital twins. Typically, photogrammetric 3D models simply take the source images as texture materials, which inherently embed unwanted lighting artifacts (at the time of capture) into the texture. Therefore, these "polluted" textures are suboptimal for a synthetic environment to enable realistic rendering. In addition, these embedded environmental lightings further bring challenges to photo-consistencies across different images that cause image-matching uncertainties. This paper presents a general image formation model for albedo recovery from typical aerial photogrammetric images under natural illuminations and derives the inverse model to resolve the albedo information through inverse rendering intrinsic image decomposition. Our approach builds on the fact that both the sun illumination and scene geometry are estimable in aerial photogrammetry, thus they can provide direct inputs for this ill-posed problem. This physics-based approach does not require additional input other than data acquired through the typical drone-based photogrammetric collection and was shown to favorably outperform existing approaches. We also demonstrate that the recovered albedo image can in turn improve typical image processing tasks in photogrammetry such as feature and dense matching, edge, and line extraction. [This work extends our prior work "A Novel Intrinsic Image Decomposition Method to Recover Albedo for Aerial Images in Photogrammetry Processing" in ISPRS Congress 2022]. The code will be made available at github.com/GDAOSU/albedo_aerial_photogrammetry

## 1. Introduction

Aerial photogrammetry nowadays has been sufficiently automated that it can (almost) generate high-resolution photorealistic models from well-collected images with a few clicks (Carroll, 2023; Fidan et al., 2023; Jian Wu et al., 2023), using commercial/open-source software packages (Agisoft, 2023; Bentley, 2022). Beyond its well-known applications to support foundational mapping, the reality-based models from photogrammetry are gaining thrusts in domains that require simulation and immersive sciences and engineering, such as virtual, augmented, extended, mixed reality (VR/AR/XR/MR), metaverse, and digital twin applications (Abdullah, 2023, 2022a, 2022b; Alidoost and Arefi, 2017; Chen et al., 2023). However, the use of photogrammetric models in these domains is still very limited, part of the reason is that the texture materials of these models are not the actual "albedos" needed by the rendering pipeline in computer graphics (Innmann et al., 2020; Lachambre, 2017). Rather, these texture materials are often directly inherited from the source images, in which the environmental lighting is unavoidably present and regarded as artifacts. For example, an albedo texture image is free of shadows so the graphics rendering pipeline can relight the model in a simulated environment, while the texture materials from source images may contain unwanted shadows that hinder the realism of the rendered views of the model. Moreover, recovering the albedos from the images may bring added benefit for necessary steps in photogrammetric processing, such as feature extraction & matching, and dense image correspondences (Song and Qin, 2022). Therefore, this not only enhances the use of photogrammetric models for extended applications but also benefits the photogrammetric process.

Albedo recovery for high-resolution images is mostly studied in the computer vision and graphics community, where given an input image, the per-pixel albedo map can be recovered through a process called inverse rendering, or intrinsic image decomposition (IID). However, among all the existing literature, albedo recovery for images of outdoor aerial mapping is less explored and remains an unsolved problem, mainly due to that capturing the complex outdoor lighting information is extremely difficult and thus it is often hard to decouple the albedo information from the shading caused by the unknown lighting (Duchêne et al., 2015; Laffont et al., 2013), such as indirect lighting, volume scattering, specular reflection.

Despite these challenges, we notice that this lighting and shading information may be partially estimated in the photogrammetry context (Chi et al., 2023): first, photogrammetric collection tasks often come with auxiliary data that contains information about the capture time and location (e.g., GPS (Global Positioning System) data), with which the solar illumination at the data capture time may be estimated. Second, photogrammetric images are often collected as aerial or oblique blocks, where reasonably accurate surface geometry can be derived to simulate shading cast by solar illumination.

Therefore, we consider the albedo recovery of photogrammetric images a tractable problem, and thus, propose a general physics-based albedo recovery approach that performs inverse rendering, or intrinsic image decomposition. The proposed method takes the above-mentioned specifics of photogrammetric images as cues to model the in-situ illumination and shading and invert the albedo from the source images. Earlier our published work (Song and Qin, 2022) presented a solution of this idea, and this work further extends it with the following contributions.

1. We present a more realistic lighting model under photogrammetric collections, which makes full use of the local geometry to model the environmental lighting.
2. We compared up-to-date data-driven methods with our albedo decomposition.

We further extend our experiments by comprehensively evaluating its scalability towards large and diverse scenes, with added experiments demonstrating the benefit of the albedo recovery for various low-level vision tasks, to inform its practical potential as part of the photogrammetric data processing pipeline.

The rest of this paper is organized as follows: **Section 2** introduces related works, and **Section 3** elaborates on the proposed outdoor lighting model consisting of directed sunlight and hemispheric skylight. **Section 4** describes our approach to estimating the proposed outdoor lighting model from a multi-view image set. In **Section 5**, we evaluated our method both quantitatively and qualitatively based on a synthetic dataset and a multi-temporal real-world UAV dataset. **Section 6** demonstrates three promising applications using our albedo imagery in fields of research and industry, and **Section 7** concludes this paper.

## 2. Related works

There have been many works in the literature that aimed to recover the albedo reflectance from single or multiple-view images in the field of Computer Vision (CV), primarily under the umbrella of Intrinsic Image Decomposition (IID), which estimates albedo, shading, and normal in the view space (Barron and Malik, 2015; Garces et al., 2022). Major efforts are to integrate learned geometric cues, or end-to-end deep learning-based estimation (Das et al., 2022; Janner et al., 2017; Li and Snavely, 2018). However, these works reflect mostly indoor scenes and with poor generalization capability. Therefore, a physics-driven method is necessary. In our context, as described in **Section 1**, the photogrammetric images indirectly provide the scene geometry with potentially estimable scene lighting, therefore adding the "physics" component into the solution. Moreover, the problem of albedo recovery is relevant to shadow removal, which presents great literature in the domain of CV, photogrammetry, and remote sensing. Considering these facts, in this section, we briefly review works related to both IID and shadow removal.

**Intrinsic image decomposition** studies the decomposition of a single image to intrinsic layers (Barrow H.G. and Tenenbaum, 1978) that include diffuse albedo (or reflectance) and shading (or pixel-level illumination). The decomposition is an ill-posed problem since both the geometry and environmental lighting are oftentimes unknown. Therefore, solutions heavily rely on regularization priors. A few early works consume user-supplied strokes to regularize the problem (Bousseau et al., 2009; Shen et al., 2011), providing low-level priors of ambient reflectance samples. Most methods assume a sparse or piecewise constant albedo (Gehler et al., 2011; Sheng et al., 2020) and smooth monochromatic illumination. Automatic methods explicitly detect shadow through supervised methods (Griffiths et al., 2022) or estimate complex shading with neural networks (Innamorati et al., 2017; Janner et al., 2017; Wang et al., 2021; Yu and Smith, 2021, 2019). These methods were proven successful for indoor scenes where the data are considered "in the same domain" as the training data. There are a few works that focus on IID for outdoor scenes, for example, by using multi-view images, Laffont et al. (2013) use a physics-based ray-tracing engine to calculate shading for reflectance recovery, while it requires in-situ measurements on site to record the environment lighting. Duchêne (2015) extracted partial cues of sky irradiance from ground images to estimate the environment irradiance, and this is partly effective since the to-be-corrected images and sky information come from the same images and thus, it does not require inter-sensor/image calibration. In an aerial/drone photogrammetry scenario, since these images do not look up to the sky, such an idea is not directly applicable. In contrast, estimated solar lighting from the position and timing (GPS and clock), which is more practical to obtain, pertains a great interest to scalable solutions like our proposed method.

**Haze removal** from outdoor images is crucial for enhancing image quality by eliminating scattered light, especially in aerial and outdoor scene analysis. Haze removal shares some common theories and techniques with albedo decomposition, though it involves different physical processes related to atmospheric and aerosol effects. The Retinex theory of color constancy (Land and McCann, 1971) is a foundational element of de-hazing algorithms (He et al., 2011; Xie et al., 2010). Recently, advancements in deep learning technology have revolutionized the field. Using neural networks to predict the Atmospheric Scattering Model (ASM) has proven efficient in decomposing haze formation and improving clarity by eliminating its effects (Cai et al., 2016; Li et al., 2017; Yang et al., 2022). A more recent trend is the end-to-end network, which implicitly encodes image enhancement by learning from a large number of hazy and haze-free image pairs to map haze to clear images. AECR-Net (Wu et al., 2021) is a compact autoencoder-like de-hazing network with contrastive regularization. Unlike most de-hazing networks that only use clear images for supervision, contrastive learning fully exploits the positive samples and negative samples during training (Chen et al., 2020). DehazeFormer (Song et al., 2023) achieves superior performance on several datasets, including ground view and remote sensing image, by adjusting the network architecture and training with large parameters. Those methods aim to remove a continuous layer of haze. In contrast, the albedo recovery focuses on surface light interaction and sometimes deals with discontinuous layers such as cast shadows.

**Shadow removal** studies the extraction and removal of cast shadows from single or multiple images. Early works focused on images with clean scene structures such as those with clean foregrounds and backgrounds, where single and isolated objects cast distinct shadows (Finlayson et al., 2004). For such a problem, simple heuristics can be applied for shadow detection followed by shadow removal. Recent studies train neural networks to perform end-to-end shadow detection and removal, which has demonstrated great success (Cun et al., 2020; Qu et al., 2017; Wang et al., 2018), but like many other deep learning approaches, its generalizability is of concern in practice. More practical solutions involve human intervention such as using strokes as guidance of the shadow region (Gong and Cosker, 2017). Regarding aerial images, shadow removal has been one of the core tasks for the photogrammetry and remote sensing community such as orthophotos production (Rahman et al., 2019; Silva et al., 2018; Zhou et al., 2021), semantic segmentation and object detection, and 3D reconstruction. One relevant line of work (Wang et al., 2017), makes use of geometry (i.e., Digital Surface Models (DSM)) to predict shaded regions using the known direction of the solar illumination, on which shadow removal can be performed using pixels on both sides of the shadow boundaries (Guo et al., 2013; Luo et al., 2019, 2018). As mentioned earlier, since the photogrammetric images indirectly provide the geometry (Duchêne, 2015; Laffont et al., 2013), our method will take advantage of such an approach, but with more comprehensive lighting modeling to recover not only shadows but general shadings caused by non-directional solar radiation (**Section 3**).

## 3. Modeling: Physics-based aerial view rendering with a general lighting model

To achieve inverse rendering, we will first formulate the aerial image rendering process with our proposed lighting model. Specifically, given the positing and timing information, we model the environmental lighting and then perform the aerial view rendering using the modeled lighting.

### 3.1 *The rendering equation*

The general rendering equation (Kajiya, 1986) models the observed radiance of surface point $p$ by a camera $L_o$, which consists of the surface-emitted radiance $L_e$ (e.g., luminescent objects) and reflected radiance from light sources $L_i$ (Equation 1). In the context of daylight optical passive sensing, $L_e$ is often not discussed since its contribution is relatively weak. The reflected radiance of a surface point is generally an integral of incident radiance $L_i$ interacted with surface BRDF (Bidirectional Reflectance Distribution Function) $f_r$ over a considered differential hemisphere $\Omega$.

$$L_o(p, \omega_o) = L_e(p, \omega_o) + \int_\Omega L_i(p, \omega_i) f_r(p, \omega_i, \omega_o)(\omega_i \cdot \boldsymbol{n})^+ d\omega_i, \tag{1}$$

where $\omega_i$ is the inlet direction of radiance from the light sources, $\omega_o$ is the outward direction from the surface to the camera, $(\cdot)^+$ is the ramp function which equivalent to $max(0, \cdot)$. At this stage, for simplicity, specular reflection will not be considered directly in the rendering equation, alternatively, if presented, they will be baked into the albedo as expected artifacts. Therefore, we will adopt the Lambertian model (Koppal, 2014) as the BRDF for our rendering model as shown in Equation 2.

$$f_r(p, \omega_i, \omega_o) = \frac{\rho}{\pi}, \tag{2}$$

where $\rho$ is albedo, which can be interpreted as the intrinsic color of the material, alternatively regarded as the reflectance in the field of remote sensing and photogrammetry. The Lambertian model describes a perfect diffusion reflection that is only affected by the incidental light but not the viewing direction (Koppal, 2014).

### 3.2 *Camera and sensor*

The rendering equation describes the physics-based modeling of light transport and its interaction with the surface materials resulting in the observed radiance. The sensor (film and its associated electronics, i.e., CCD or CMOS) of the camera records the observed radiances and processes them into machine-readable digital signals (pixel values) through an A/D (Analog / Digital) converting process. On top of this process, various standard internal processes were conducted by the camera, including tune mapping, white balancing, and sometimes compression. This facilitates a perceptually well-balanced image for visualization, however, may add complexities when interpreting radiance out of it. As a result, image pixel values may not be linearly correlated to the scene radiance (Grossberg and Nayar, 2004), thus it is challenging to model. Luckily, most aerial and drone-based cameras allow the export of raw images at the time of collection, which produces raw pixel values with minimal internal processing. Therefore, raw pixel value intensity $I$ can be assumed the following linear relationship with the radiance $L_o$ (Equation 3), where $\epsilon$ is a scale factor that can be interpreted as the exposure factor.

$$I = \epsilon L_o. \tag{3}$$

### 3.3 *Modeling outdoor illumination*

Outdoor illumination consists of both directional illumination from solar radiation and ambient illumination from the sky (scattered illumination, hereafter called sky illumination). This interprets the inlet lighting $L_i$ as a compounded source from both Sun $L_i^{sun}$ and Sky illumination $L_i^{sky}$ (Equation 4), time-varying lights based on solar radiation (with a time variable $t$).

$$L_i(p, \omega_i, t) = L_i^{sun}(p, t) + L_i^{sky}(p, \omega_i, t), \tag{4}$$

where $p$ is a point on the surface, $\omega_i$ is the direction of the incident light, $t$ is time. It should be noted that indirect lights, such as lighting reflected from other objects, will not be considered in the aerial case since they are ignorable as stated in our earlier work (Song and Qin, 2022).

*Modeling Sun Radiance*

Despite the Sun illumination should be theoretically modeled as a point/area light source, due to its far distance to Earth, the solid angle of the sun from the Earth can be as small as 0.68×10⁻⁴ steradians (Wald, 2018). Thus, it is mostly assumed to be parallel and directional light to a region of interest. Therefore, the inlet sunlight to a surface point $p$ can be effectively modeled in the following (Equation 5):

$$L_i^{sun}(p,t) = \psi^{sun}(p,t) \cdot V^{sun}(p, \omega_{sun}(t)) \,, \tag{5}$$

where $\omega_{sun}$ refers to the sun angle as a function of time $t$, which represents the lighting direction. $V^{sun}$ refers to the visibility of the sun when considering a surface point $p$ (determined by the local geometry of the scene and sunlight direction), as it can be possibly occluded (e.g., shadows). $\psi^{sun}$ refers to the source intensity of the sun, a function of its incident angle (determined by the location $p$ on the surface), and data collection time. In the context of albedo correction, $\psi^{sun}$ is represented as $\psi^{sun} \in \mathbb{R}^3$ (RGB) to consistent with the image. In practice, the intensity of $\psi^{sun}$ can be defined relative to unit 1, referring to the largest magnitude of a year. Therefore, given the surface point $p$, local geometry (determined by photogrammetric 3D reconstruction), and the time (determining the sunlight direction and strength), $L_i^{sun}$ is directly calculable.

*Modeling Sky Radiance*

When traveling through the atmosphere, the sunlight can be scattered through the aerosol and various layers, creating a domed light source, with centers around the sunlight direction. An extreme example of such is a cloudy day, where direct sunlight and strong shadows are not observable. Following the definition of a dome light, we model the skylight as the following (Equation 6):

$$L_i^{sky}(p, \omega_i, t) = \psi^{sky} \cdot G(\omega_i - \omega_{sun}(t)) \cdot V^{sky}(p, \omega_i) \,, \tag{6}$$

where $V^{sky}$ is skylight visibility at a surface point $p$ (determined by the local geometry of the scene) observing the direction $\omega_i$. $\psi^{sky}$ is the source intensity of the sky, which is considered a constant value by assuming a uniform skylight. $G(\omega_i - \omega_{sun}(t))$ refers to a Gaussian function (with a controllable, but large variance) indicating the maximal intensity still occurs at the solar illumination direction.

Different from our earlier work (Song and Qin, 2022), this paper combined skylight visibility (as shown in Figure 1) and uniform source intensity to create a non-homogenous skylight, closer to the heterogeneity nature of such a lighting model.

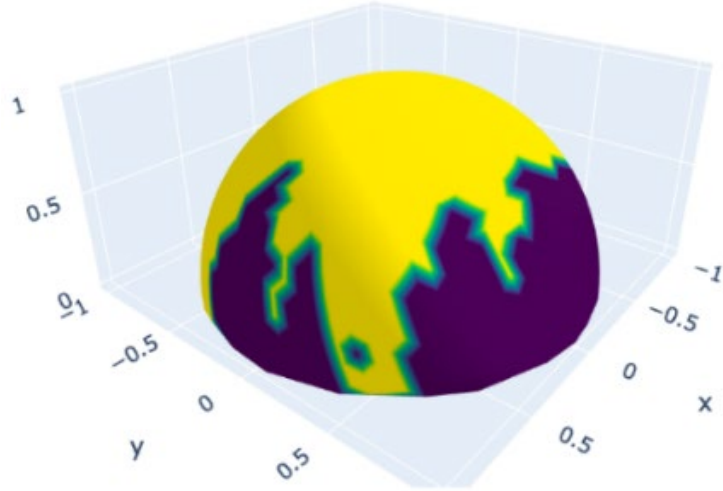

Figure 1. Example of $V^{sky}$ from a surface point (center of the hemisphere).

### 3.4 Connecting sun and sky illumination

As described in **Section 3.3**, with photogrammetrically reconstructed 3D geometry, $L_i^{sun}$ is calculable given a surface location and time. However, $L_i^{sky}$, as shown in Equation 6, is not directly calculable using the same information, primarily due to that the light intensity constant $\psi^{sky}$ is unknown. Considering that this quantity still origins from the sunlight ($\psi^{sun}(p,t)$), our aim is to seek for the relationship between them. We noted a critical fact: in an ideal case, the shadowed region, due to occlusion, contains no sunlight $\psi^{sun}(p,t)$, but it will still be illuminated by the skylight, since the skylight is a domed light that comes from all directions. Whereas non-shadowed region contains both sunlight and skylight. This implies that by using pixel intensity values between shadowed and non-shadowed regions, it is possible to build the relations among these two quantities ($\psi^{sun}$ and $\psi^{sky}$), thus resolving $L_i^{sky}$ as a function of known $\psi^{sun}$. The pixel intensity values in shadowed and non-shadowed regions can be easily observed using lit-shadow pixel pairs, where a pair of pixels span between shadowed and non-shadowed regions, such shown in Figure 2, can be found. A similar idea using lit-shadow subtraction has been applied to estimate atmosphere optical depth with in Mars orbiter images (Hoekzema et al., 2011). It should be noted that the distance of two pixels $p$ and $p + \Delta$ is the best minimal to remove other factors, such as non-homogenous camera sensor responses, as well as the change of sky visibilities $V^{sky}$ due to distant locations. Land and McCann (1971) proposed to assume albedo ρ across the scene should be mostly with low frequency, i.e., piecewise linear or constant. Therefore, taking this assumption, we can reasonably assume that these two points, since they are close enough, share the same albedo ρ. At the same time, we assume that the visibility of the sky $V^{sky}$ of these two points are the same: because the shadow and non-shadow regions only reflect the visibility difference of light from a single direction (the same as the direct sunlight), while $V^{sky}$ considers accumulated impacts of all directions, thus differences by a single light direction can be ignored.

These facts can be implemented by building the following observational constraints between lit-shadow points (Equation 7).

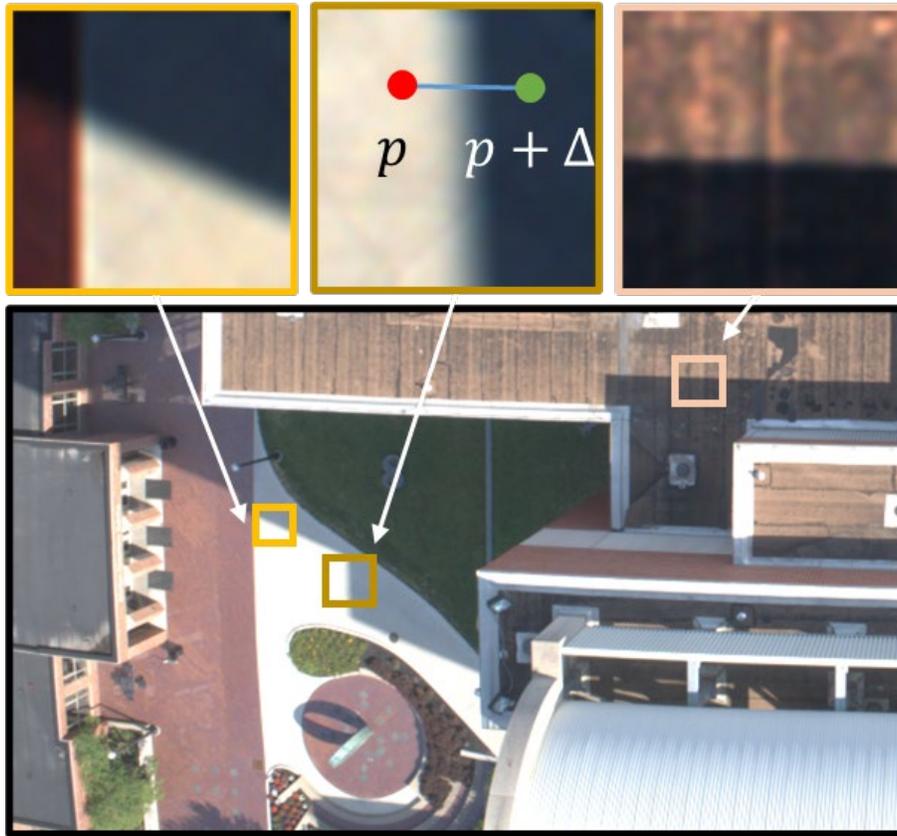

Figure 2. Lit-Shadow pair near the shadow boundary. The figure shows 3 patches containing casted shadows with penumbra width (transitions between shadow to non-shadow region).

$$\begin{cases} \rho_{lit} & = & \rho_{shadow} = \rho \\ n_{lit} & = & n_{shadow} = n \\ V_{lit}^{sky} & = & V_{shadow}^{sky} = V^{sky} \\ V_{lit}^{sun} & = & 1 \\ V_{shadow}^{sun} & = & 0 \end{cases} \quad . \tag{7}$$

Then, we expand Equation 4 for these two lit-shadow points, which leads to the following (Equation 8). Note for simplicity, we ignored the variable $t$, since these two points (in the same image) are collected at the same time:

$$\begin{cases} L_o(p) = & \rho\psi^{sun}(p)(\omega_{sun}\cdot n)^+ + \rho\psi^{sky}\int_\Omega V^{sky}(p,\omega_i)G(\omega_i-\omega_{sun}(t))(\omega_i\cdot n)^+d\omega_i \\ L_o(p+\Delta) = & \rho\psi^{sky}\int_\Omega V^{sky}(p,\omega_i)G(\omega_i-\omega_{sun}(t))(\omega_i\cdot n)^+d\omega_i \end{cases}, \quad (8)$$

where the sunlight and skylight shading, as part of Equation 8, can be denoted as $S^{sun}$ and $S^{sky}$ (Equation 9):

$$\begin{cases} S^{sun} := (\omega_{sun}\cdot n)^+ \\ S^{sky} := \int_\Omega V^{sky}(p,\omega_i)G(\omega_i-\omega_{sun}(t))(\omega_i\cdot n)^+d\omega_i \end{cases}. \quad (9)$$

Here, we assume another simplification: given that the sky visibility is minimally impacted by local geometry, we can assume its visibility component $\int_\Omega V^{sky}(p,\omega_i)d\omega_i$ is close to full visibility (i.e., 1), while the cumulated incident angle $\int_\Omega G(\omega_i-\omega_{sun}(t))(\omega_i\cdot n)^+d\omega_i$ is biased towards the sunlight incident angle $(\omega_{sun}\cdot n)^+$. This leads to the following conclusion (Equation 10): where the skylight intensity $\psi^{sky}$ and sunlight $\psi^{sun}$ is up to a constant factor calculable using lit-shadow pairs (detectable, introduced in **Section 4.2**).

$$\phi := \frac{\psi^{sky}}{\psi^{sun}} = \frac{L_o(p+\Delta)}{L_o(p)-L_o(p+\Delta)}. \quad (10)$$

As a result, given a surface location $p$, local geometry (from photogrammetry), and collection time $t$, both the $L_i^{sun}$ and $L_i^{sky}$ are calculable.

## 4. Solution: Inverse rendering for albedo recovery

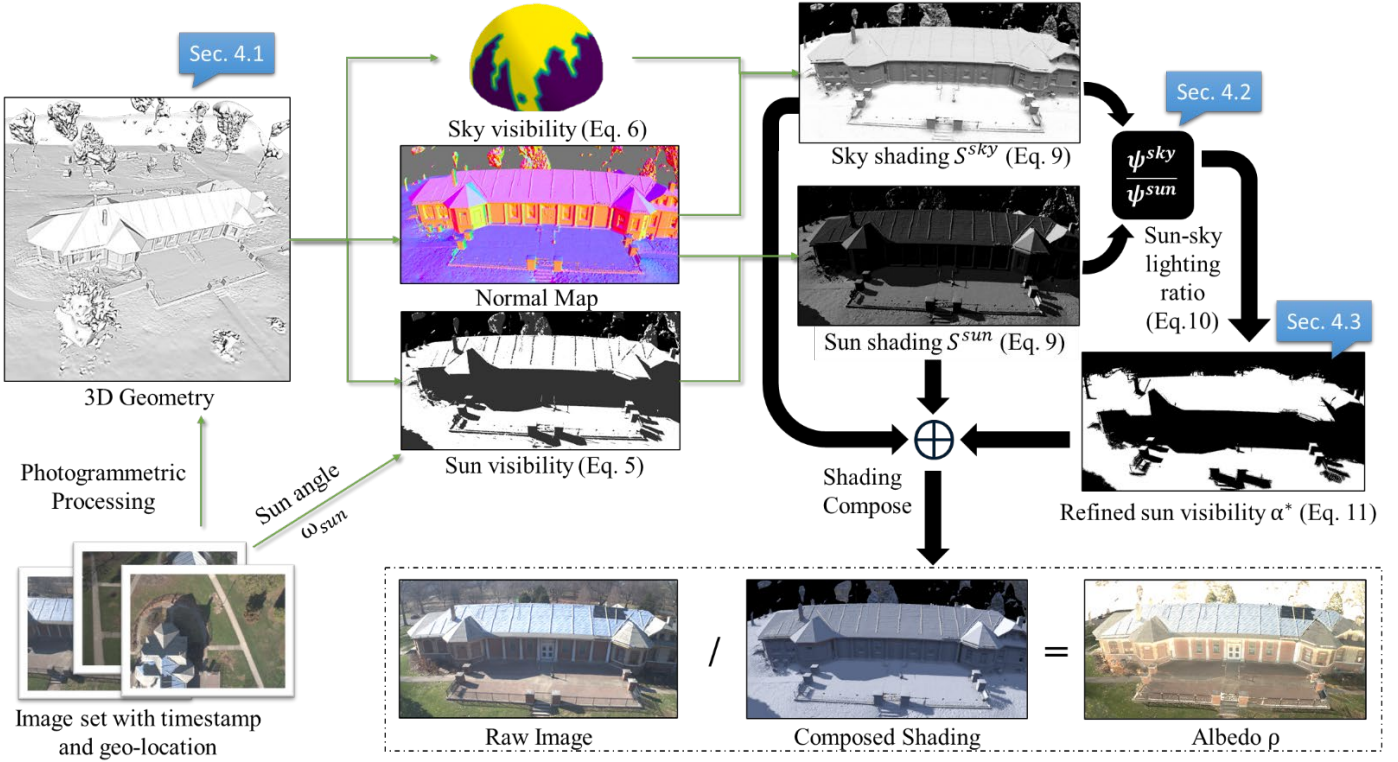

Figure 3. The workflow of our solution in albedo recovery.

Based on **Section 3**, with photogrammetric images and their associated location and collection time, the inlet source lights, consisting of the direct sunlight $L_i^{sun}$ and skylight $L_i^{sky}$, are calculable. Therefore, based on the general rendering equation (Equation 1), the emitted light $L_e(p,\omega_o)$ ignored under our context), since $L_i(p,\omega_i)$ is calculable, as well as the geometry (normal $n$, derived from photogrammetry), the albedo (denoted as the BRDF $f_r$) can be easily inverted. Hence, our solution built on this under the context of photogrammetric images can be depicted as a workflow shown in Figure 3. First, we perform standard photogrammetric data processing to calculate the pose of the images and generate the 3D meshes of the scene. It should be noted that for data collection, we require the users to store the raw images, as well as the metadata including the GPS location and the time of the data collection, to calculate the inlet light. Second, we prepare these metadata to resolve the Sun visibility for both the sunlight and skylight ($V^{sun}$ and $V^{sky}$), which is used to compute the sunlight and skylight ($L_i^{sun}$ and $L_i^{sky}$). Third, the $L_i^{sky}$ is calibrated in reference to $L_i^{sun}$ is done by using a lit-shadow pairs introduced in **Section 3.4**. Fourth, the shadow predicted by Sun visibility $V^{sun}$ is often a binary mask, which cannot depict the penumbra effects (transition between shadow and non-shadow, shown in Figure 2), hence we propose to refine the $V^{sun}$ as a soft value (between 0-1 instead binary) to cope with this effect (to be introduced in **Section 4.3**), and finally optimize the recovered albedo ρ.

**4.1** *Photogrammetric data preparation*

We captured a regular photogrammetric block as the input and ran through a standard photogrammetric data processing pipeline to orient images and generate the high-quality meshes using off-the-shelf commercial or open-source photogrammetric software such as Bentley ContextCapture (Bentley, 2022). As described in **Section 3.2**, our albedo recovery method requires the pixel color intensity to be best linearly correlated with the radiance, we store the RAW images. If RAW images are not available, color space calibration algorithms can be performed pre- or post-flight (Lin et al., 2004; Tai et al., 2013). Our proposed method assumes input from well-collected photogrammetric models, from which the data derived from photogrammetry preserves adequate depth and surface normal for most pixels. Typically, dense matching algorithms in the photogrammetric process enforce a smooth constraint, which may cause suppression of high-frequency details of the geometry for small and complex objects. However, since the loss of the details only takes a small and sparse set of pixels, the erroneous surface normal would unlikely cause substantial changes, as the rendering equation only takes the cosine of the normal, which is relatively robust to directional changes of a sparse set of surface normal. In the event the 3D models are reconstructed by poorly collected images, the lack of geometry accuracy may lead to more substantial errors in albedo recovery.

**4.2** *Calculating sunlight and skylight*

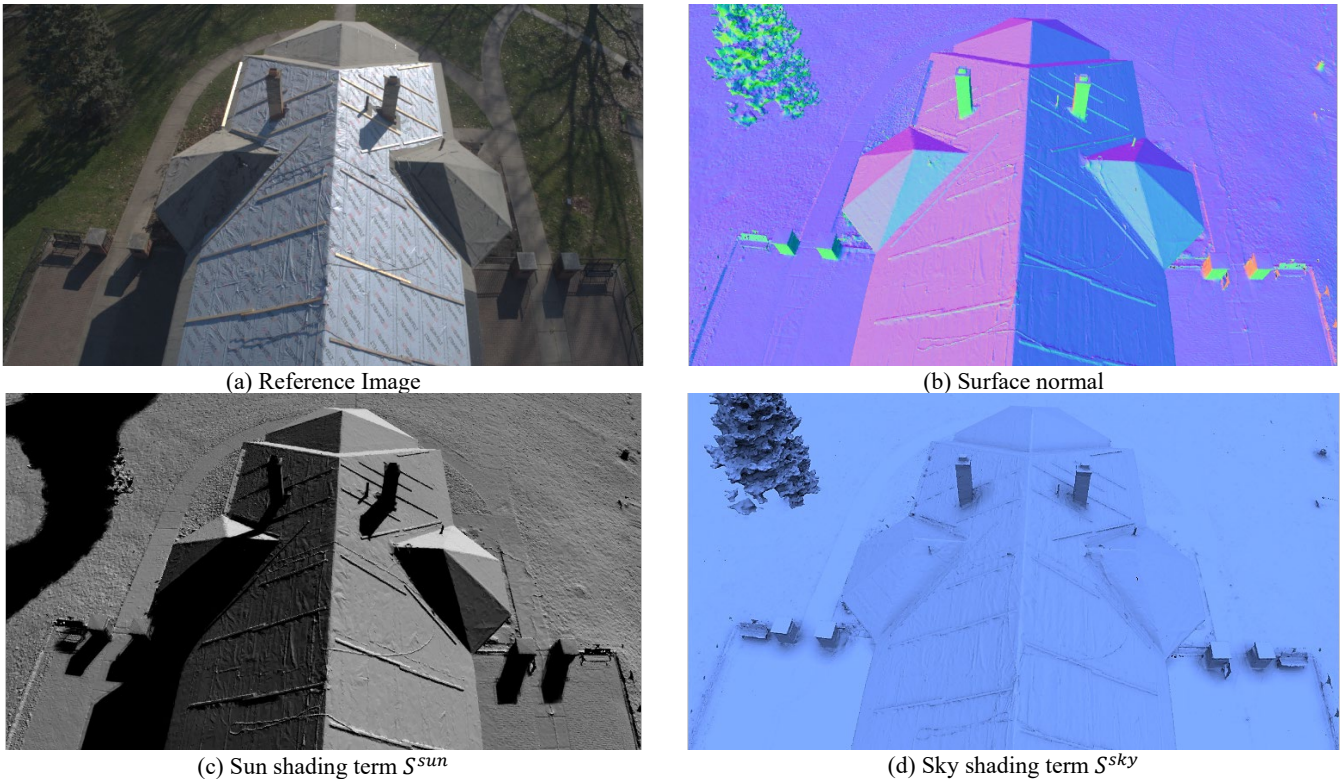

(a) Reference Image  (b) Surface normal

(c) Sun shading term $S^{sun}$  (d) Sky shading term $S^{sky}$

Figure 4. Visualize shading components of sunlight and skylight.

*Computing lighting components*

Using meta information, the sun position $\omega_{sun}$ can be easily approximated by the astronomical almanac's algorithm (Michalsky, 1988) with known geolocation and date time. For each view, we project rays from the camera center and apply the path-tracing technique (Wald et al., 2014) to detect their intersecting point to the surface for depth computation and occlusion detection. Surface normal can then be computed from the depth (**Error! Reference source not found.**b). From the depth image, sun visibility $V^{sun}$ can be computed by emitting rays from every single pixel to the sunlight direction $\omega_{sun}$, and performing occlusion direction. This allows to compute the shading of both sunlight and the skylight ($S^{sun}$ and $S^{sky}$, Equation 9), where $S^{sun}$ is easily computed by taking the sunlight direction, and $S^{sky}$ is computed by a sample of 1024 points over the hemisphere. Results of the shading components are visualized in Figure 4(c-d), which perceptually matches the intuition of shadings caused by direct sunlight and that caused by cloudy-day light.

*Resolving Sun-sky lighting ratio Φ with lit-shadow pairs*

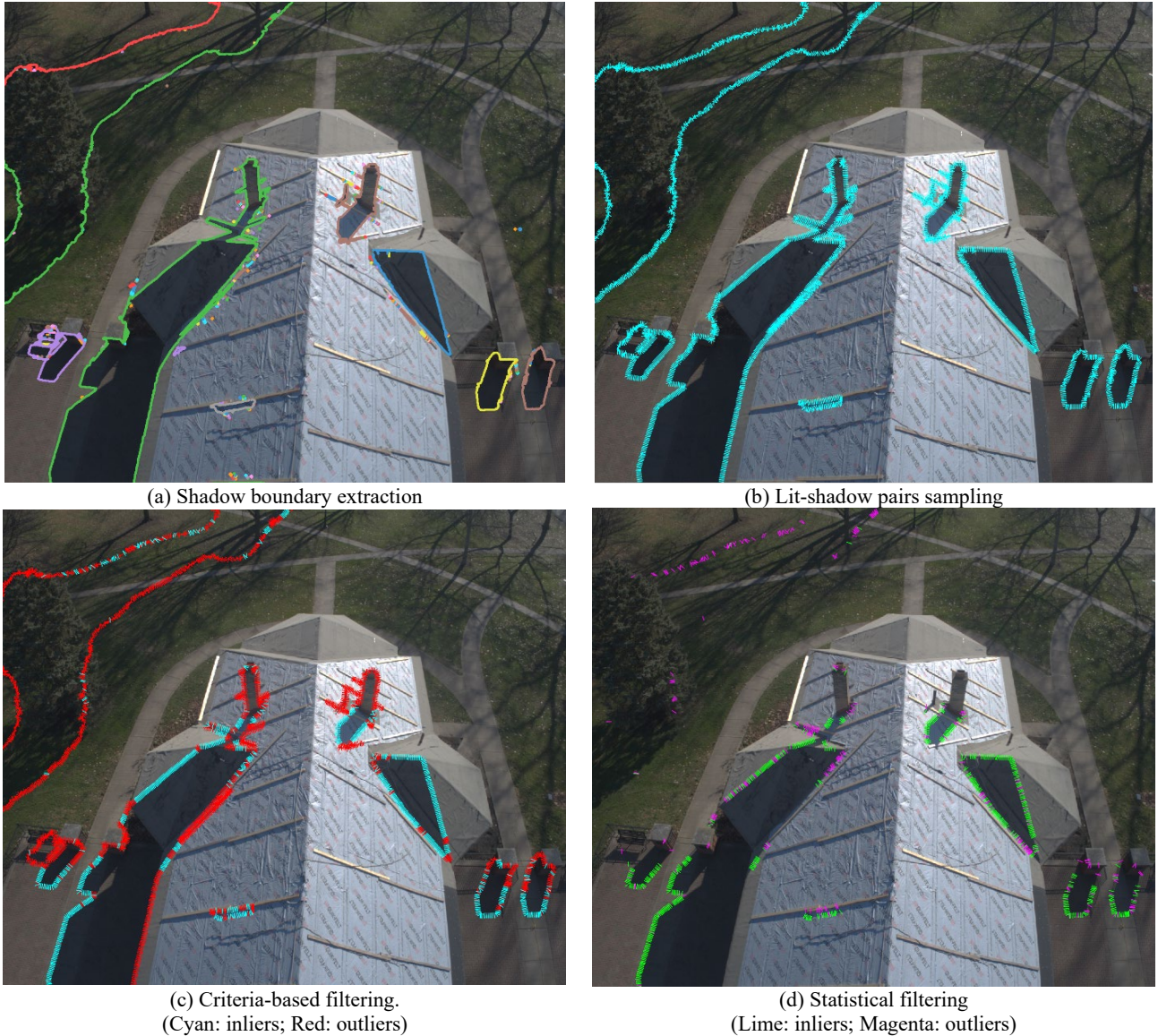

(a) Shadow boundary extraction  
(b) Lit-shadow pairs sampling  
(c) Criteria-based filtering. (Cyan: inliers; Red: outliers)  
(d) Statistical filtering (Lime: inliers; Magenta: outliers)  
Figure 5. Find reliable lit-shadow pairs to estimate Φ.

As described in **Section 3.4**, connecting the intensity of sunlight and skylight requires pixel values of paired points sitting in shadow and non-shadow neighbors (following Equation 10). To detect such paired points, we propose a filtering strategy to sample, and then filter pairs for building robust lit-shadow statistics. Firstly, given the sun's visibility $V^{sun}$, we extract the boundaries of shadow regions (Figure 5(a)) and sample pairs in the shadow and non-shadow regions (Figure 5(b)). Secondly, we propose a set of criteria to filter out unwanted pairs that are potentially outliers, i.e., those that do not share similar albedo values. The criteria are listed in **Algorithm** 1 to form a criteria-based filter using both pixel intensity and geometry (depth continuity) as intuitive measures, for example, pixels with under or overexposure should not be considered, and pixels with large depth difference should not be considered (since they may not lie on the same surface). Figure 5b and Figure 5c show results before and after applying the filtering. To ensure these lit-shadow pairs provide a robust estimation of $\Phi$, we calculate $\Phi$ for each of the filtered pairs and fit them to a Gaussian distribution. With a p-test, if the null hypothesis is not rejected (meaning the p-value is smaller than 0.05 to achieve a 95% confidence level of fitting), we will compute the mean of the samples within the 95% data (adjusted mean) as the final ratio $\Phi$. Figure 5d shows outliers identified as in the 5% tail of the distribution. The rest of the lit-shadow pairs are reliable sources for the $\Phi$ estimation.

**4.3** *Sun visibility refinement to cope with the Penumbra effect*

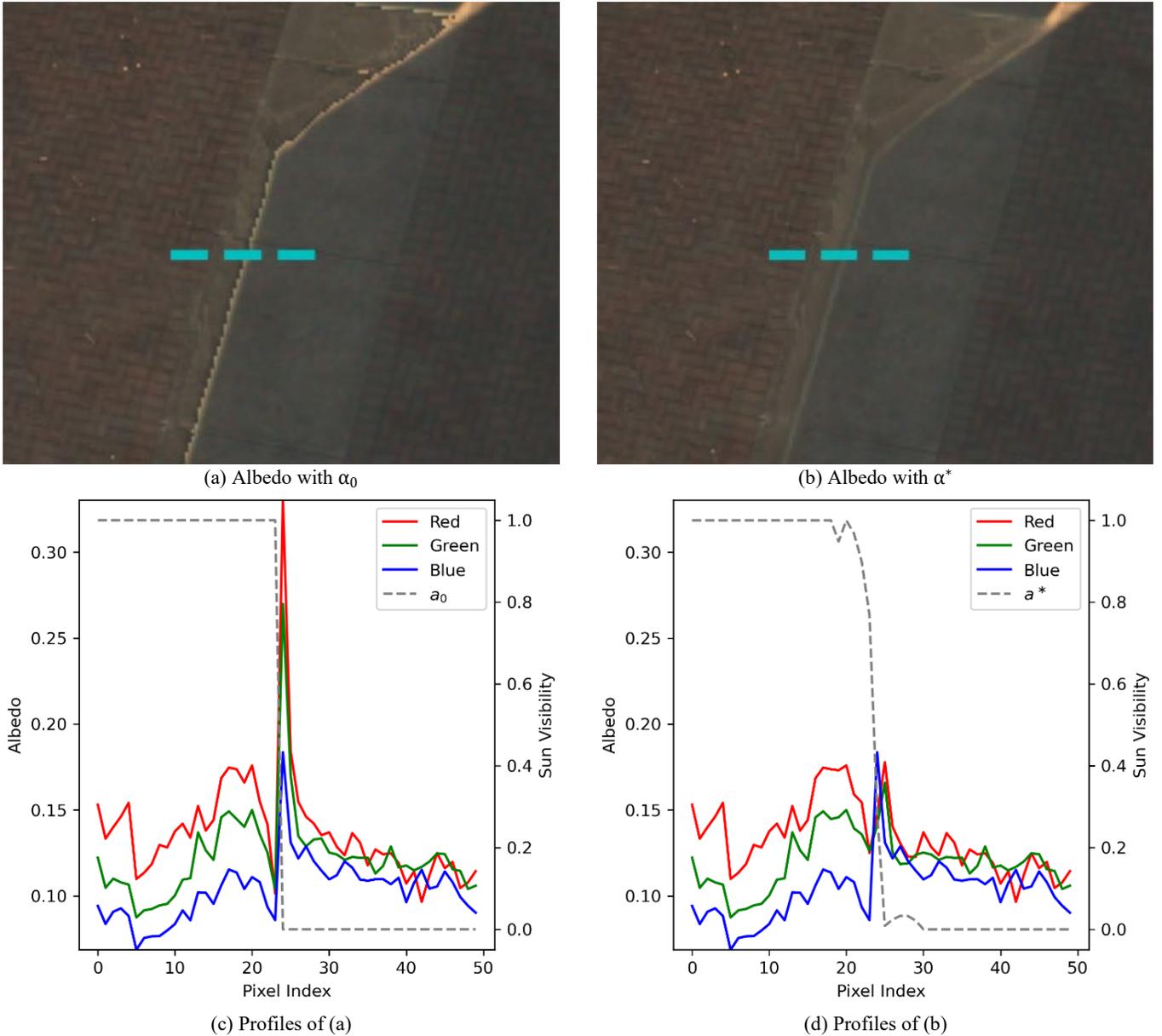

(a) Albedo with $\alpha_0$
(b) Albedo with $\alpha^*$
(c) Profiles of (a)
(d) Profiles of (b)

Figure 6. (a) and (c) are the recovered albedo with binary $\alpha_0$ showing artifacts in both the figure and the profile; (b) and (d) show the effectiveness of the recovered albedo using our Sun visibility refinement.

As mentioned at the beginning of **Section 4**, once the inlet light is known and the geometry is known, the albedo $\rho$ can easily be recovered by inverting the rendering equation by using $\rho = L_o/(\psi^{sun}S^{sun} + \psi^{sky}S^{sky})$. However, since we assume the sun visibility $V^{sun}$, i.e., the shadow, as a binary variable, does not match the actual penumbra effect reflecting the physics of the smooth transition between shadow and non-shadowed region due to sun disc scattering. If not refining the $V^{sun}$, the albedo recovery will produce artifacts at the shadow boundaries (as shown in Figure 6a). an analysis of the pixel intensity profile shown in Figure 6(c) indicates that the discontinuity of the visibility profile produced such an artifact. To address this, we aim to recover a continuous $V^{sun}$ using a total variation (TV) regularization, as shown in Equation 11:

$$\begin{aligned}\alpha^*(x) &= \arg\min_{\alpha(x)} ||\alpha(x)-\alpha_0(x)||_P^2 + \left|\nabla \frac{1}{\rho(x)}\right|, \\ \frac{1}{\rho(x)} &= \frac{\psi^{sun}S^{sun}(x)}{L_o(x)}\alpha(x) + \frac{\psi^{sky}S^{sky}(x)}{L_o(x)} \\ &= \psi^{sun}(\frac{S^{sun}(x)}{L_o(x)}\alpha(x) + \frac{\Phi \cdot S^{sky}(x)}{L_o(x)}),\end{aligned} \quad (11)$$

where α is the sun visibility along the profile, $x$ is the distance on the profile, $\alpha_0$ is the initial binary sun visibility from the directional sun model, $\|\cdot\|_P$ is Mahalanobis distance with weight matrix $P$, $\nabla$ is Gradient operator. We choose to minimize 1/ρ since compared with directly optimizing regarding ρ, the 1/ρ yields a closed-form solution due to its linearity with $\alpha(x)$. Weight matrix $P$ is a diagonal positive definite matrix. This formulation adjusts the $\alpha(x)$ that close to the shadow boundary, and $L_o(x)$ in the formulation inherently uses the image information to guide the sun visibility refinement. By optimizing based on Equation 11, the sun visibility becomes a continuous variable, and the produced artifacts can be successfully removed (Figure 6(b) and (d)). In our approach, we assume the $\psi^{sun}$ to be a constant vector since only the ratio between sky and sun intensity can be estimated $\Phi = \psi^{sky}/\psi^{sun}$ (Equation 10).

Combining $\alpha^*$ and $\Phi$ into the lighting formulation, the inverted albedo Figure 7 (b) presents a more consistent color on areas of the same material compared to the original image shown in Figure 7 (a). The composed shading is shown in Figure 7 (c).

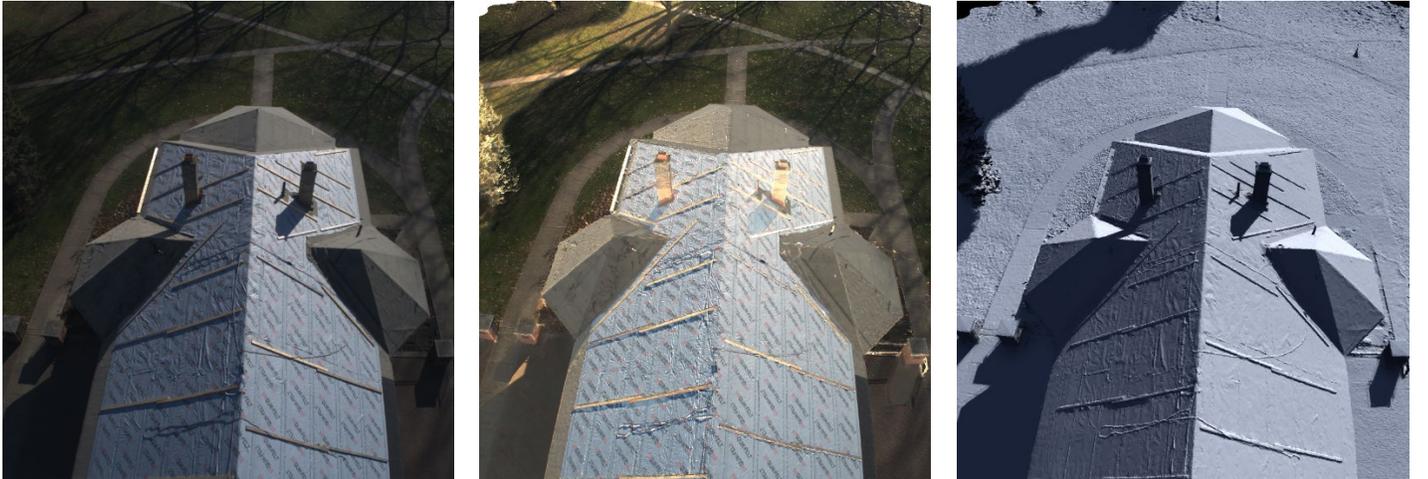

(a) Input image  (b) Our estimated albedo image  (c) Our estimated shading image
Figure 7. Our albedo and shading estimation.

## 5. Evaluation

In this section, we conduct albedo recovery experiments using both synthetic datasets and real-world datasets to evaluate our method. Since there exists no ground-truth albedo for real-world outdoor images, we only qualitatively and indirectly evaluate the results. Quantitative results are concluded using simulated results through physics-based rendering (PBR). We also compare our results with those of the state-of-the-art methods including InverseRenderNet (IRN) (Yu and Smith, 2019), Shadow Matting Generative Adversarial Network (SMGAN) (Cun et al., 2020), InverseRenderNetv2 (IRNv2) (Yu and Smith, 2021), Weakly-Supervised Single-View Image Relighting (WSR) (Yi et al., 2023), Intrinsic Image Diffusion (IID) (Kocsis et al., 2024), and our 2022 version (Song and Qin, 2022). A comprehensive evaluation protocol for real-world images is adopted to evaluate: 1) if the recovered albedo is free from cast shadow and shading effects (**Section 5.3**); 2) if the recovered albedo shows consistency for images of the same outdoor scene collected at different time of day (**Section 5.4**).

## 5.1 *Dataset for experiments*

No publicly available dataset provides high-quality 3d models, and outdoor aerial views as long as the corresponding ground-truth albedo. Thus, in our experiment, we created a synthetic dataset for the quantitative evaluation. To demonstrate the effectiveness of our method with real-world data, we also collected aerial images with a drone under different lighting conditions. The details of each type of dataset are presented in the following content.

### *Synthetic dataset*

To create a synthetic city model to simulate aerial photogrammetry, we generate a synthetic 3D city model using ESRI CityEngine, a procedural modeling software (ESRI, 2020). The model textures come from a pre-existing asset library without any baked shadows or shading. Blender Cycles (Blender Online Community, 2021) is a ray-tracing render engine that allows us to generate photorealistic samples with ground truth albedo and lighting components. We use built-in Nishita sunlight and skylight models (Nishita et al., 1993) to simulate the natural lighting environment. To evaluate the performance of our method in cities of different landscapes, we created two scenes featuring high-rise buildings and low-rise buildings respectively. Our synthetic dataset includes 30 virtual camera images for each scene, ground truth albedo images, camera orientation parameters, and sun positions. All images use linear RGB color space and are stored in OpenEXR (Academy Software Foundation, 2023) format for the wide dynamic range. The average ground resolution is about 20 cm/pixel ~ 50 cm/pixel. Figure 8 presents an example of our dataset.

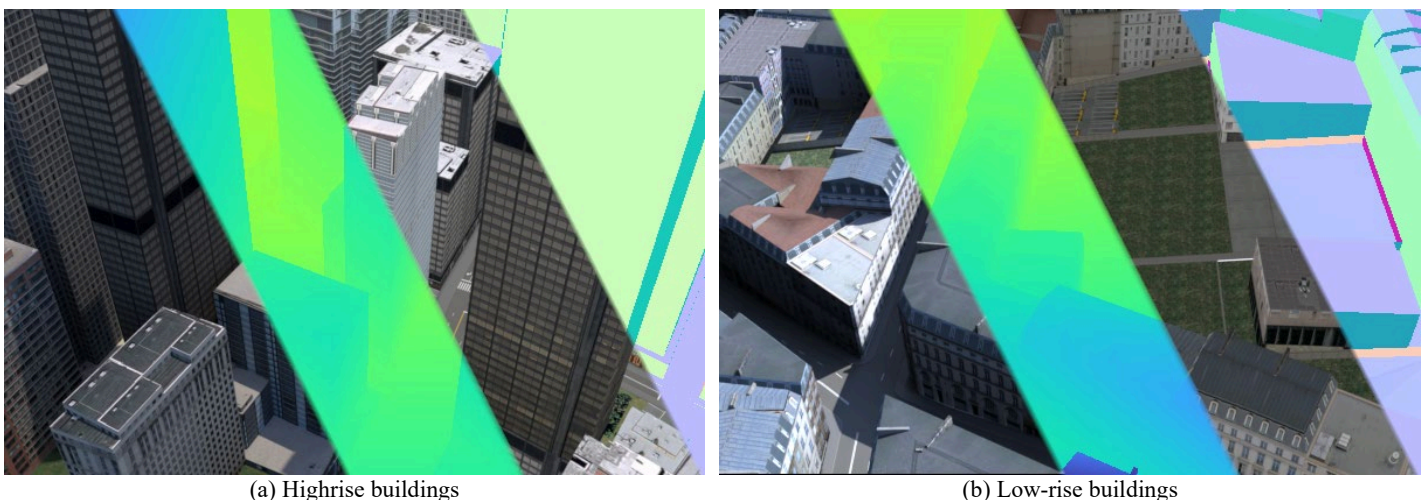

(a) Highrise buildings  (b) Low-rise buildings

Figure 8. Our synthetic dataset features both high-rise and low-rise buildings. From left to right in each image, we display 1) rendered image, 2) depth map, 3) diffuse albedo, and 4) surface normal.

### *Real-world aerial dataset*

Capturing ground-truth albedo for all regions in a scene in the wild is extremely challenging (Jiaye Wu et al., 2023). Thus, we collect drone photogrammetric blocks over three days focusing on the same scene. The goal is to apply our albedo recovery method to those individual images and evaluate cross-view consistency under different lighting. We captured aerial images using a DJI FC6310S camera with an 8.8 mm f/2.8 lens. The camera was carried by a DJI Phantom Pro 4 v2.0. Flight height is 70 meters above ground, and average ground resolution is 2.5 cm/pixel. The region of data collection is around a squared area of 200 × 200m. We had 6 flights on 3 consequential days to collect images under various illumination conditions (two-time points per day), as listed in Table 1. All flights are taken on the same site but at different times of the day. Day-1 data includes nadir images Day-2 and Day-3 data include oblique images as well.

Table 1. Data acquisition time and number of images of each flight.

| | | |
|---|---|---|
| **Day 1** | 08:00 AM – 09:00 AM \| 29 images | 10:00 AM – 11:00 AM \| 31 images |
| **Day 2** | 09:00 AM – 10:00 AM \| 52 images | 06:00 PM – 07: 00 PM \| 53 images |
| **Day 3** | 09:00 AM – 11:00 AM \| 68 images | 11:00 AM – 12:00 PM \| 57 images |

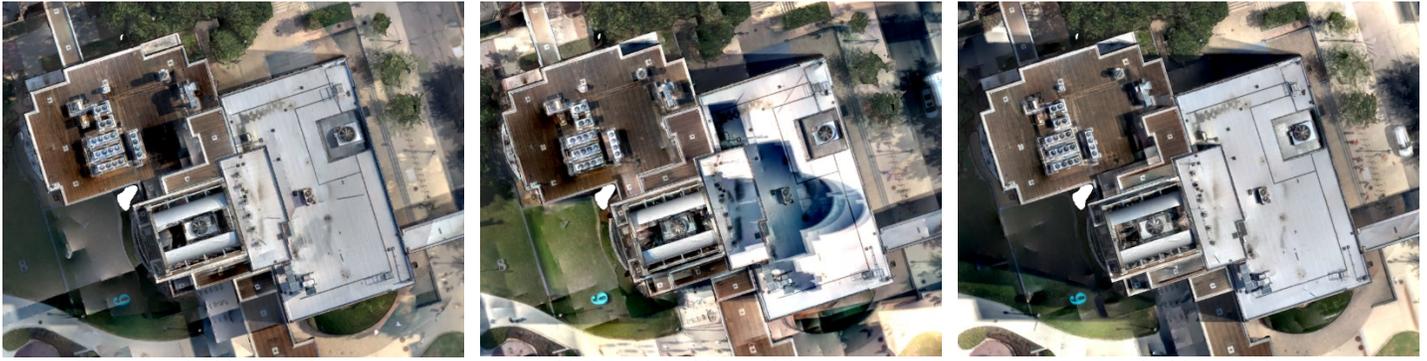

(a) Textured with Day 1 images.  (b) Textured with Day 2 images.  (c) Textured with Day 3 images.
Figure 9. Orthorectified imagery from the original images of our multitemporal dataset. The images were taken two times in one day (see Table 1), resulting in significant overlapping cast shadow artifacts.

We perform photogrammetric 3D reconstruction of the data on Day 3, since all images are taken in the morning and show reflect better geometric reconstruction. Then, we registered Day 1 and Day 2 images with virtual Ground Control Points (GCP) manually collected from the model of Day 3, with registration errors smaller than 0.01m. As shown in

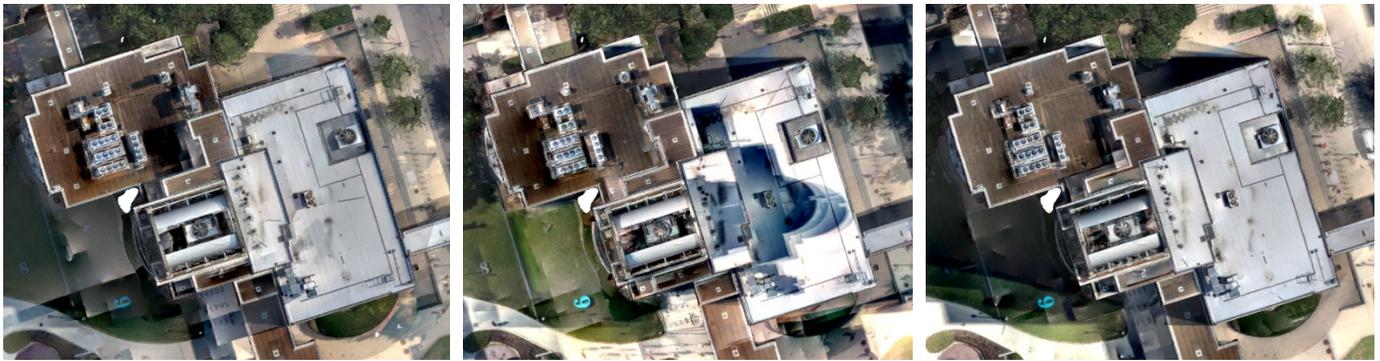

(a) Textured with Day 1 images.  (b) Textured with Day 2 images.  (c) Textured with Day 3 images.

Figure 9, we generated an orthorectified imagery of each day using a common geometry reconstructed from Day 3 images.

### 5.2 *Quantitative evaluation with the synthetic dataset*

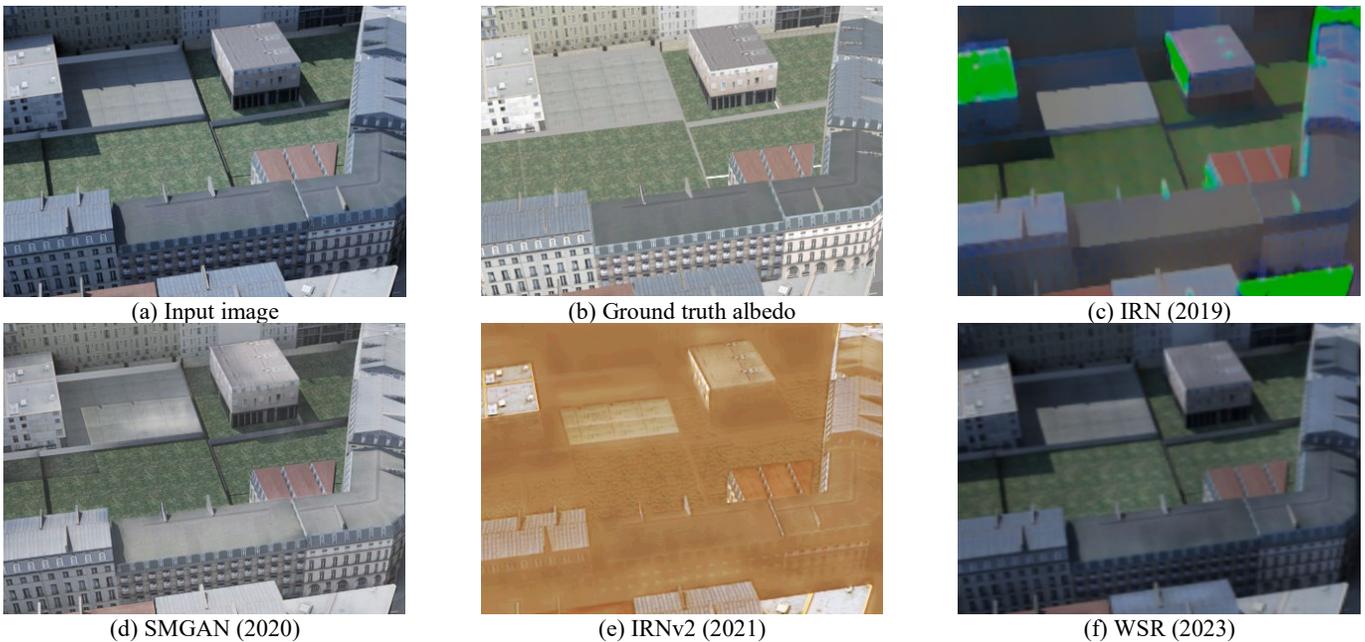

(a) Input image  (b) Ground truth albedo  (c) IRN (2019)
(d) SMGAN (2020)  (e) IRNv2 (2021)  (f) WSR (2023)

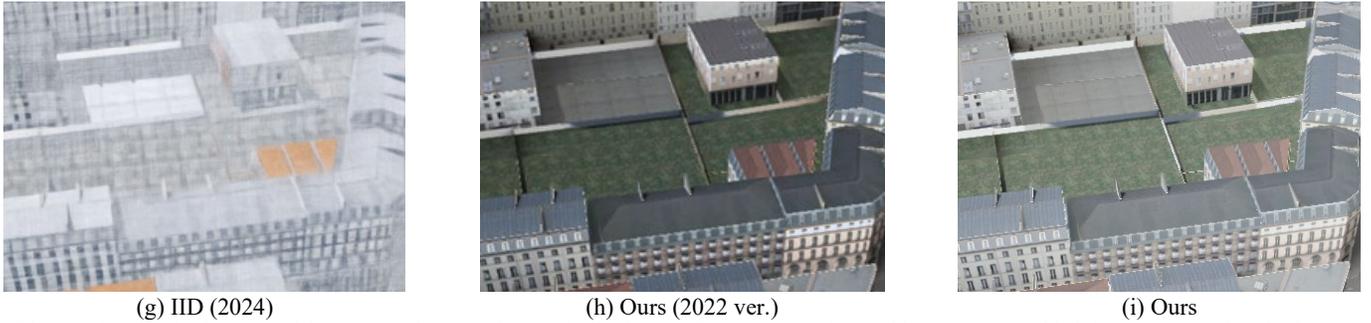

| (g) IID (2024) | (h) Ours (2022 ver.) | (i) Ours |

Figure 10. Albedo decomposition comparison on the synthetic dataset. Images are enhanced by contrast and brightness correction for better visualization.

We compare our albedo decomposition method to data-driven methods. Since they are not finetuned with the synthetic dataset we used in this paper, the resolved albedo could be an up-to-scale value across different methods. Thus, we adopt scale-invariant MSE (SMSE) (Grosse et al., 2009; Yi et al., 2023) and local scale-invariant MSE (LMSE) (Grosse et al., 2009) as evaluation metrics. Besides those, we also introduced DreamSim (Fu et al., 2023), a learned perceptual metric to evaluate the albedo decomposition. As shown in Table 2, our framework outperforms all learning-based methods and has comparable performance with our previous method proposed in (Song and Qin, 2022). It can be seen that our earlier version of the approach marginally outperforms the proposed approach in the high-rise building region. This is due to that high-rise and dense buildings may hold stronger indirect lighting effects, i.e., scattering light bouncing among multiple buildings. While both approaches do not account for indirect shading, the earlier version with a simpler skylight mode may be less affected and presents a better "averaging" effect over unmodeled errors. However, considering that the differences are marginal, the new sky light model still yields better performance. The baseline is evaluated with the rendered image to the albedo. Among the comparing approaches in the LMSE metric, only SMGAN shows positive recovery of albedo (better than the baseline) in the high-rise building region. We visualize the recovered albedo with all methods in Figure 10.

### 5.3 Qualitative evaluation with real-world dataset

We evaluate the performance of our albedo decomposition on real-world aerial datasets by comparing it with existing albedo recovery or related algorithms. Figure 11 shows the visual quality of the recovered albedo: Among the results of different approaches, IRN tended to recover an over-smooth albedo image, while IRNv2 preserved more details, but the albedo image was significantly distorted (both color and geometry). The typical qualitative evaluation focuses on the rectangle region in Figure 11(c, h), where it is expected that shadows, and over-bright concrete/pedestrian ways (due to direct sunlight) should be corrected. We observe that our method produces the best performance in shadow removal, correcting the over-brightness of concretes. Among the other comparing methods, SMGAN obtained notable results, while it produced artifacts at the rightmost rectangle.

### 5.4 Multi-temporal consistency with real-world datasets evaluation

We evaluate the recovered albedo of datasets of the same scene but collected on different days (**Section 5.1**). The goal is to apply our albedo recovery method to these individual images taken at different times and evaluate their consistency. Ideally, the recovered albedo will appear the same, albeit their original images are significant. Figure 12 shows an example of our result: Figure 12(a)-(b) shows two original images taken at different local time, where drastically different illuminations can be found, with our albedo correction method, as shown in Figure 12(c)-(d), it shows much less shading effect and higher consistency.

Table 2. Comparison of albedo decomposition results with ground truth. The Baseline is evaluated with the rendered images without any process.

| Methods | High-rise buildings | | | Low-rise buildings | | | Average | | |
|---|---|---|---|---|---|---|---|---|---|
| | SMSE↓ | LMSE↓ | DreamSim↓ | SMSE↓ | LMSE↓ | DreamSim↓ | SMSE↓ | LMSE↓ | DreamSim↓ |
| Baseline | 0.02641 | 3.09944 | 0.164947 | 0.05345 | 10.3575 | 0.501852 | 0.03993 | 6.72847 | 0.333400 |
| IRN (2019) | 0.02618 | 4.92464 | 0.496914 | 0.03724 | 8.20697 | 0.570274 | 0.03171 | 6.56580 | 0.533594 |
| SMGAN (2020) | 0.01972 | 2.58756 | 0.145072 | 0.02912 | 4.51363 | 0.180541 | 0.02442 | 3.55059 | 0.162807 |
| IRNv2 (2021) | 0.02408 | 4.48446 | 0.341791 | 0.01918 | 3.59038 | 0.356102 | 0.02163 | 4.03742 | 0.348947 |
| WSR (2023) | 0.02603 | 3.61332 | 0.187022 | 0.02665 | 4.31144 | 0.252010 | 0.02634 | 3.96238 | 0.219516 |
| IID (2024) | 0.02536 | 4.26920 | 0.293189 | 0.02331 | 4.21016 | 0.446355 | 0.02433 | 4.23968 | 0.369772 |
| Ours (2022 ver.) | **0.01097** | **2.23774** | **0.093060** | 0.02179 | 3.41106 | 0.128822 | 0.01638 | 2.82440 | **0.110941** |
| Ours | 0.01295 | 2.26250 | 0.117938 | **0.01719** | **3.24758** | **0.114001** | **0.01507** | **2.75504** | 0.115970 |

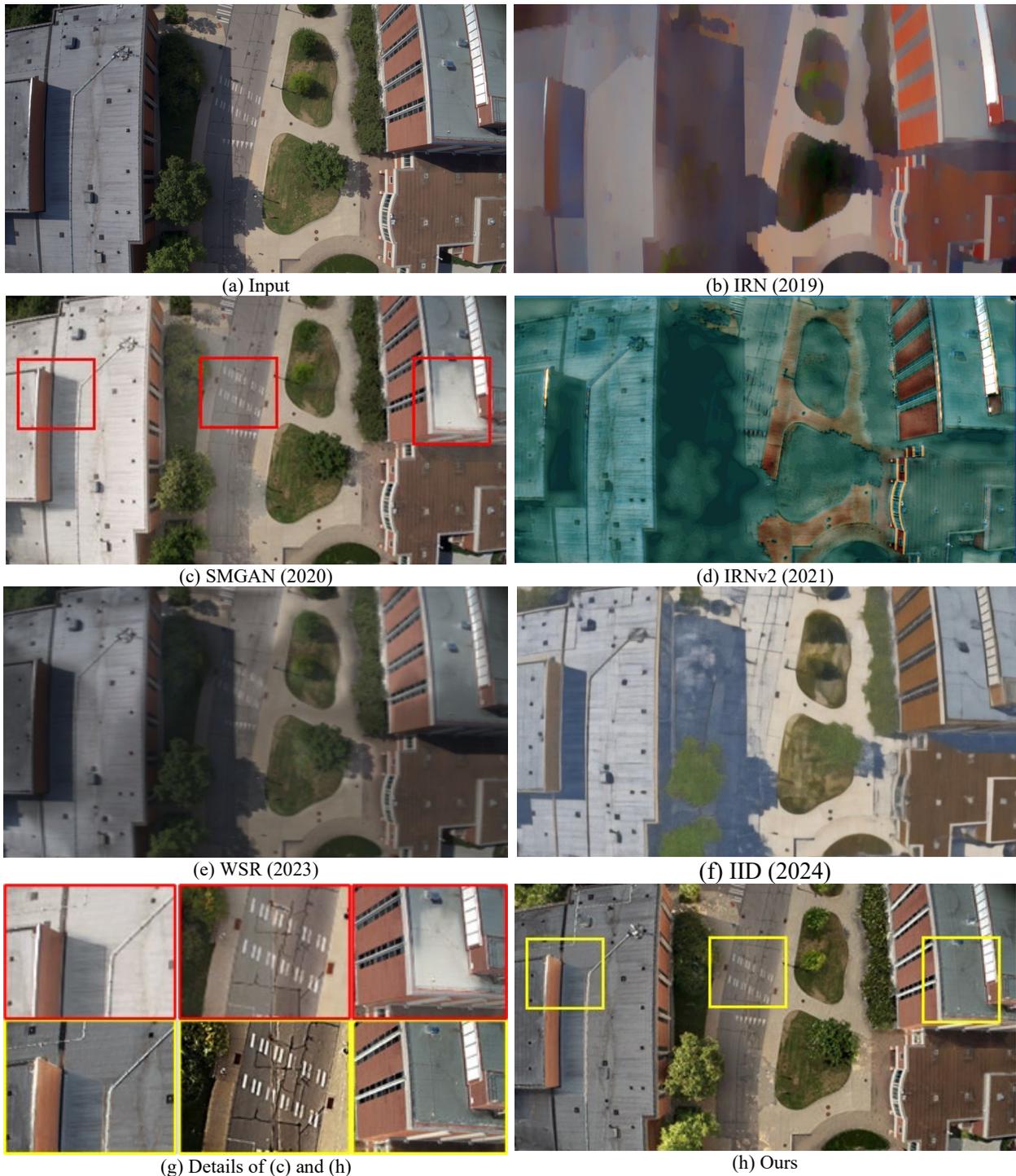

Figure 11. Albedo decomposition comparison on a real-world dataset. Images are enhanced by contrast and brightness correction for better visualization.

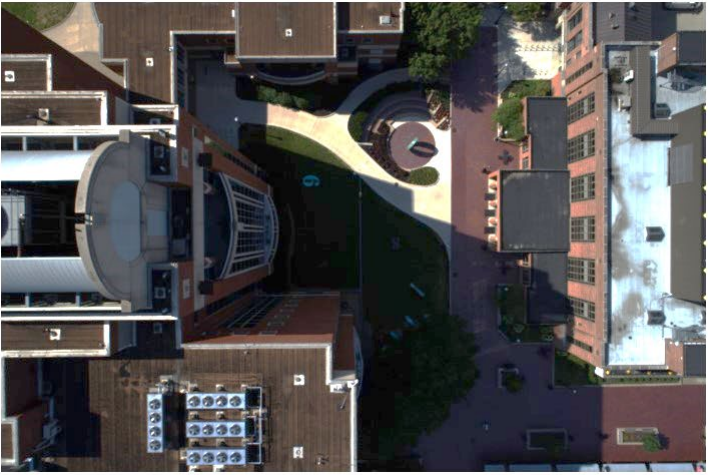 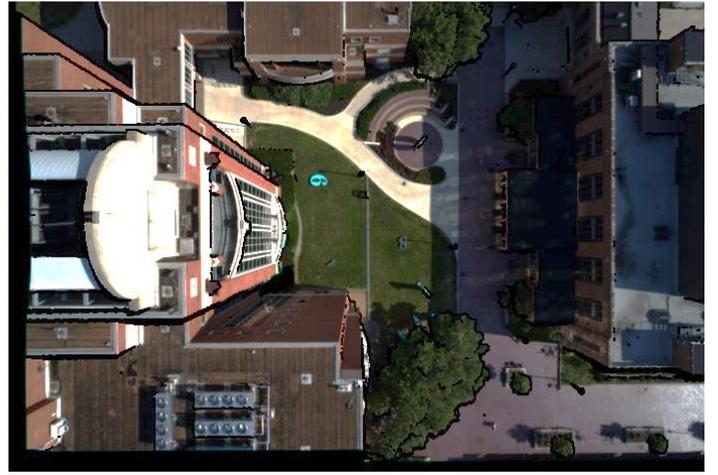

(a) Original image taken at 10 AM local time　　　　　　　　(b) Original image taken at 6 PM local time

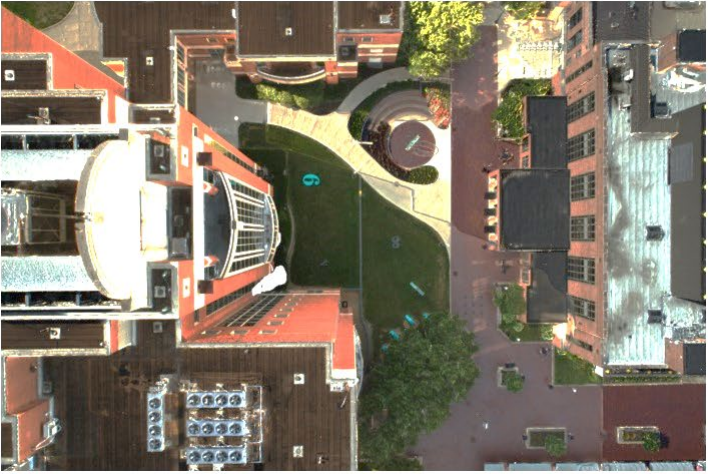 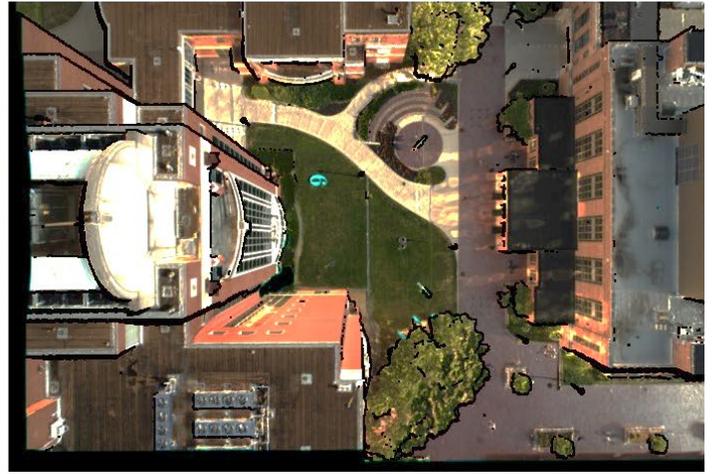

(c) Our albedo recovered image of (a)　　　　　　　　(d) Our albedo recovered image of (b)

Figure 12. Evaluate the temporal consistency of our albedo decomposition (c)-(d) compared with original images (a)-(b). All images are geometrically corrected to the same viewpoint. Black regions in (b) and (d) are due to occlusions from view correction.

To quantitatively evaluate this, we composite mean images from multi-temporal images, then compute the error between multi-temporal images to the mean images, with and without albedo recovery (i.e., original and albedo recovered image). To facilitate pixel-wise computation, images of different dates and times are geometrically corrected to have the same viewpoint (using the photogrammetric mesh). As expected, the temporal consistency of our albedo decomposition significantly excels the consistency of the original images. Table 3 shows that with a scale of 8-bit grey scale, the albedo images of different days and times are 32% more consistent than the original images in terms of standard deviation. It should be noted that the statistics in Table 3 are derived from all 6 flights collected across 3 days (as shown in Table 1).

Table 3. Distribution of errors of multi-temporal images (Digital Numbers, scaled in 0-255).

|  | Original images | Our albedo images | Error drops ↓ |
|---|---|---|---|
| **25th Abs. Err.** | 2.33 | 2.00 | -14.16 |
| **Median Abs. Err.** | 14.67 | 10.67 | -27.26 |
| **75th Abs. Err.** | 49.33 | 31.33 | -36.48 |
| **Max Abs. Err.** | 168.67 | 169.67 | 0.592 |
| **Standard Deviation** | 23.69 | 15.8 | -33.30 |

## 6. Applications

To demonstrate the possibility of utilizing our approach in fields of research and industry, we present four applications benefiting from our albedo recovery method: 1) Model relighting: the model textured with albedo images is more realistic in simulation system relighting; 2) Feature point extraction: our albedo recovered image yield more feature point matches (sparse features) for photogrammetric processing; 3) Dense matching: our albedo recovered image produce more complicate dense matching results in stereo and multi-view reconstruction. 4) Change detection: our albedo recovered image drastically improves the change detection application at ultra-high-resolution images.

## 6.1 *Application: textured model relighting*

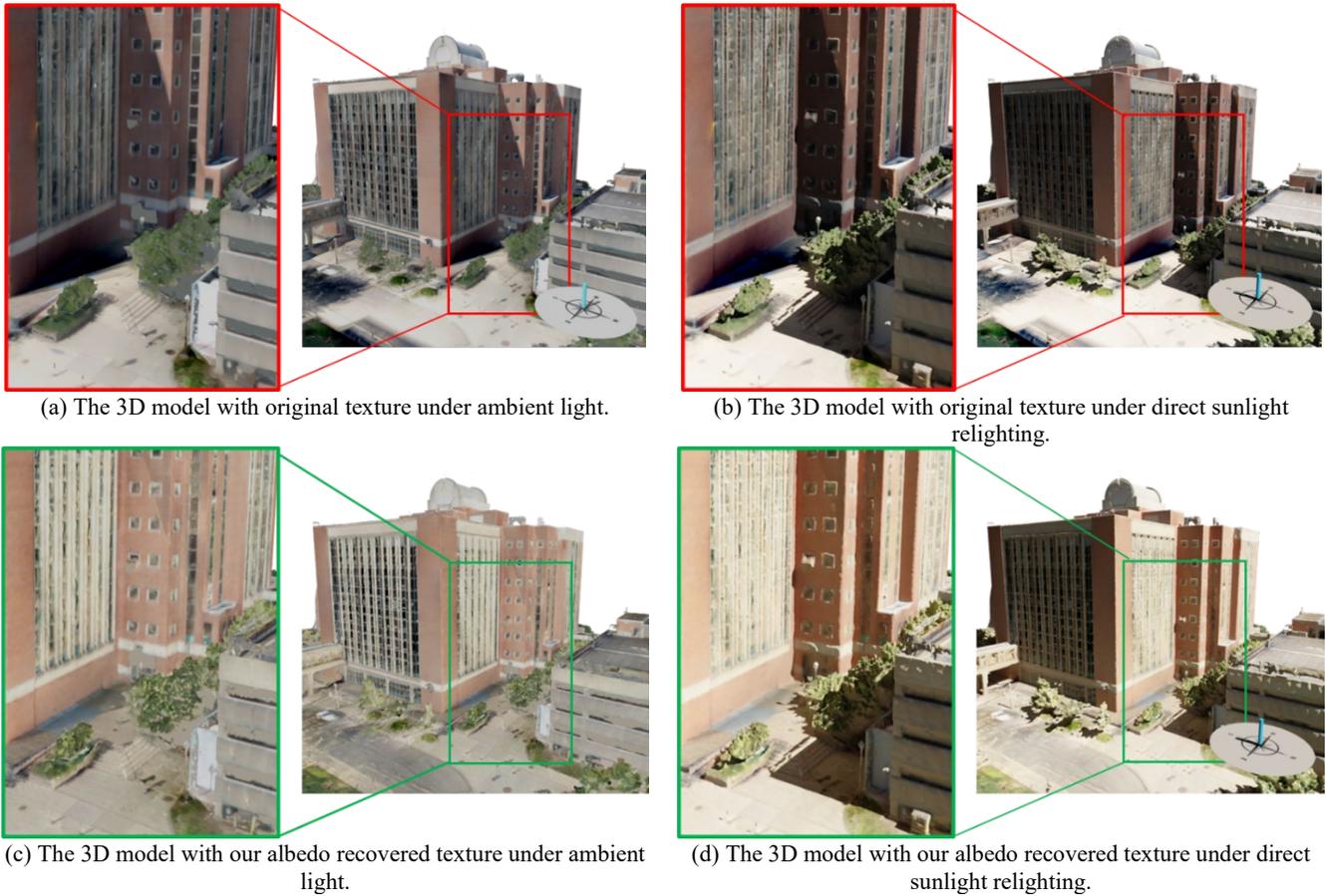

(a) The 3D model with original texture under ambient light.

(b) The 3D model with original texture under direct sunlight relighting.

(c) The 3D model with our albedo recovered texture under ambient light.

(d) The 3D model with our albedo recovered texture under direct sunlight relighting.

Figure 13. Rendered textured model from novel view with ambient lighting (a, c) and simulated sun-sky lighting (b, d). The compass and cast shadow on the right corner indicate the sun position in image capturing or rendering.

Model relighting is a standard application of 3D textured models in a simulation system, in which views are rendered under different simulated lighting, and thus the realism of the rendered view is the key. In Figure 13, we show that, with our recovered image, the rendered views contain much fewer shading artifacts. Figure 13 (a) (c) are rendered with ambient lighting (area and homogenous lighting) and Figure 13 (b) (d) are rendered with a different sun position than that at the collection time. Ideally, under ambient lighting, no shadow or shading effect should be observed. With sunlight, the shadow azimuth and shading should be coherent with physical law. As can be seen in Figure 13, the rendered view using the model texture of the original images contains unwanted shadows under the ambient light and double shadows under the sunlight. Our model shows a much better rendered view since there are nearly no shading artifacts, and colors are consistent for materials that are supposed to be consistent (e.g., the paved road materials).

## 6.2 *Application: feature extraction & matching, and edge detection*

In feature matching, ideally, more consistent images yield more feature matches, because the photo-consistency of difference images is a key factor of concern that impacts the performance of many feature extractors and matches (Braeger and Foroosh, 2021; Valgren and Lilienthal, 2010). Therefore, we compare the performance of feature point matching on a pair of images before and after our albedo correction. Figure 14 shows the performance of a classic feature matching, SIFT (Lowe, 2004), as it is still the most widely used feature matcher in photogrammetry software packages. We show that the performance of the handmade feature matcher, when applied to our albedo recovered image, significantly outperforms its results on the original image, in terms of the distribution of matches, as well as the number of inliers.

Recent advances in deep learning (DL) based matching show promising results with images under diverse illumination, viewing angles, and scales. We applied a state-of-the-art matching approach using the SuperPoint (Sarlin et al., 2020) descriptor with LightGlue (Lindenberger et al., 2023) on images before and after albedo decomposition. Figure 15 presents the matching results at full resolution. Compared to the classic matcher, the DL-based method finds more inliers and performs similarly on images before and after albedo correction. Unlike the classic matcher, the DL-based method is more robust to cast shadows and changes in illumination. The number of inliers is similar before and after albedo correction. However, we noted that the DL-based method produces a larger mean distance to epipolar lines (y-parallax) than SIFT, indicating that the quality of the keypoint location is not as good as the classic method.

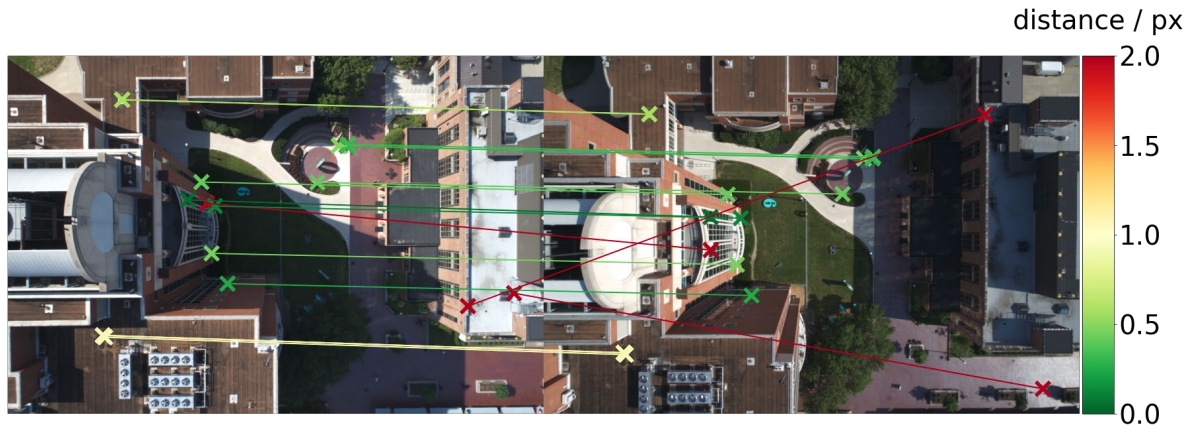

(a) Original image pair
(Number of candidates: 4929, number of Inliers with RANSAC: 17, mean distance to epipolar lines: 35.42 pixels)

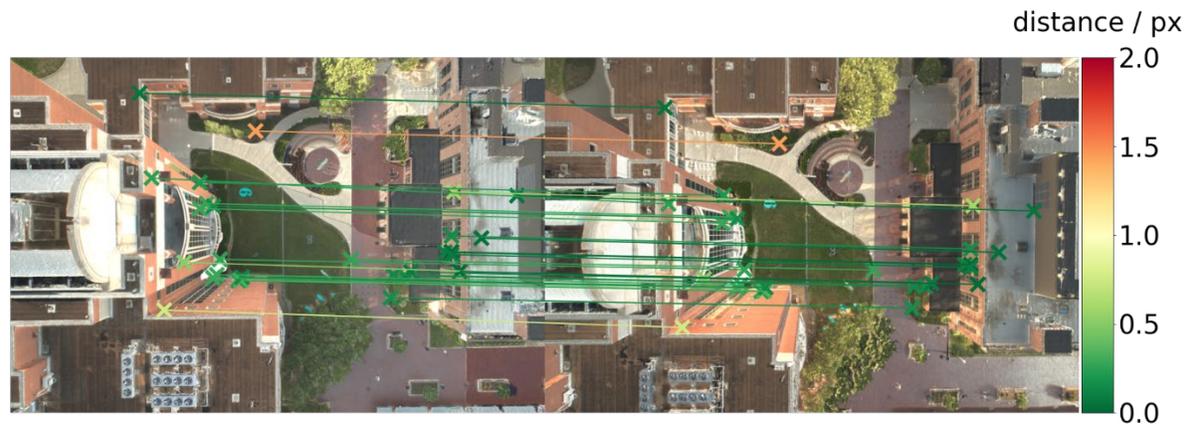

(b) Our albedo image pair
(Number of candidates: 8328, number of Inliers with RANSAC: 28, mean distance to epipolar lines: 0.26 pixels)

Figure 14. SIFT matching across different times of the day. Matching candidates are generated using a Brute-Force matcher. Inliers are filtered with the Lowe's ratio test (Lowe, 2004) (less than 0.8) and fundamental matrix estimation using RANSAC (reprojection threshold less than 0.1 pixel). Lines and points are colored by distance to corresponding epipolar lines (y-parallax) where the Fundamental matrix is computed from camera poses from the structure-from-motion with all images.

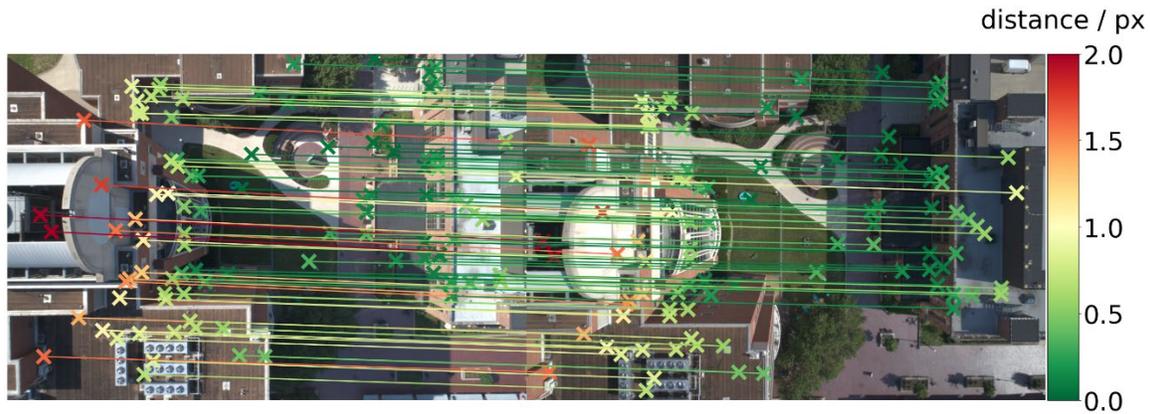

(a) Original image pair
(Number of candidates: 1470, number of Inliers with RANSAC: 94, mean distance to epipolar lines: 0.55 pixels)

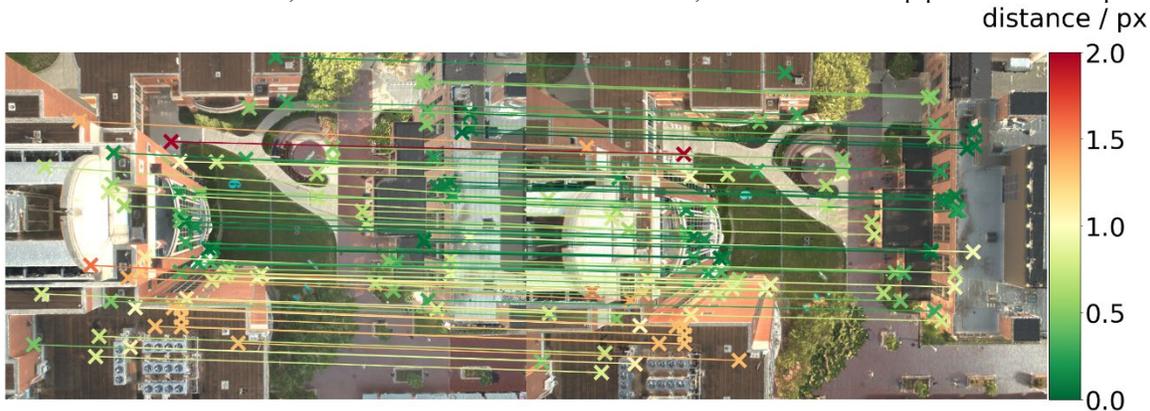

(b) Our albedo image pair
(Number of candidates: 1335, number of Inliers with RANSAC: 79, mean distance to epipolar lines: 0.6189 pixels)

Figure 15. SuperPoint/LightGlue (Lindenberger et al., 2023) matching across different times of the day. Matching candidates are generated using a Brute-Force matcher. Inliers are then filtered through fundamental matrix estimation with RANSAC, applying a reprojection threshold of less than 0.1 pixel. Lines and points are colored by distance to corresponding epipolar lines (y-parallax) where the Fundamental matrix is computed from camera poses from the structure-from-motion with all images.

Edge and line extraction is an important task that serves for feature matching and reconstruction, which is sensitive to shading and shadows as well. Thus, we compare the performance of Canny edge detection (Canny, 1986) and Line Segment Detector (LSD) (Grompone von Gioi et al., 2012) on the original and our albedo decomposition image. As shown in Figure 16, both canny edges and line segments are less affected by cast shadow, and there are more features under the shadowed region.

The recent advanced data-driven method performs line description and matching using a neural network. We applied the cutting-edge line segment matching algorithm, SOLD2 (Pautrat et al., 2021), to evaluate the effectiveness of our albedo decomposition. As shown in Figure 17, the original image and the albedo recovered image were processed for line detection and matching. The matched colors indicate the correspondence of line segments between the two images. It can be observed that more matches were found in the image pair without the cast shadow (Figure 17b) compared to the original images, especially in areas covered by the cast shadow.

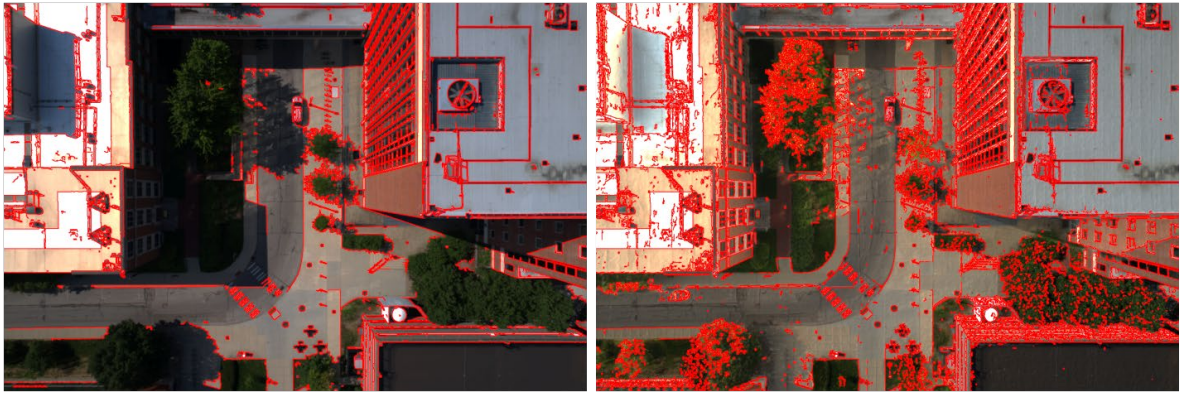

(a) Canny edge of the original image
(b) Canny edge of our albedo decomposition

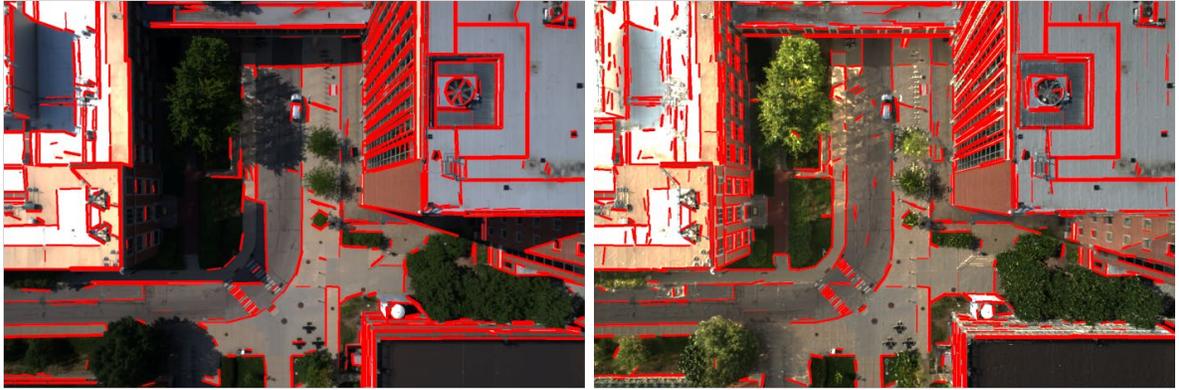

(c) LSD segments of the original image
(d) LSD segments of our albedo decomposition

Figure 16. Edge and line segment extraction using the original image and our albedo decomposition.

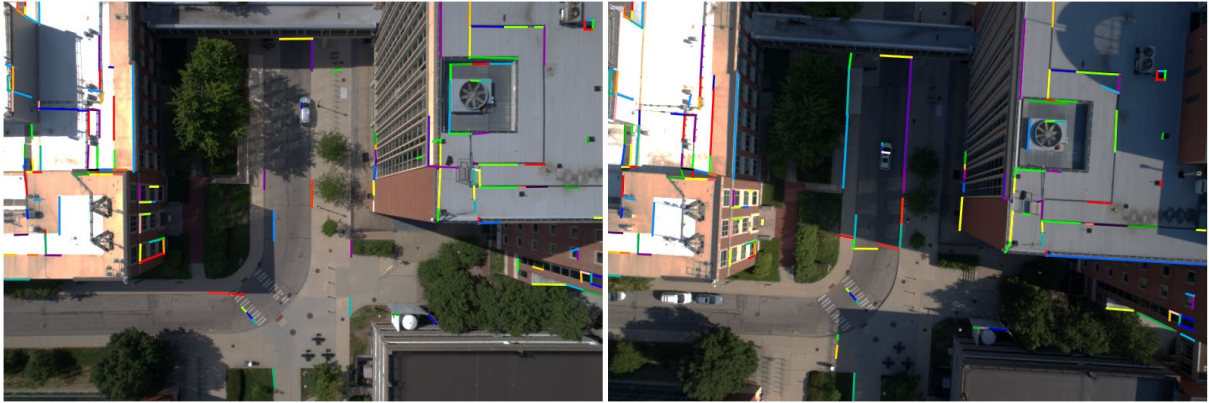

(a) Matched line segments of the original image pair

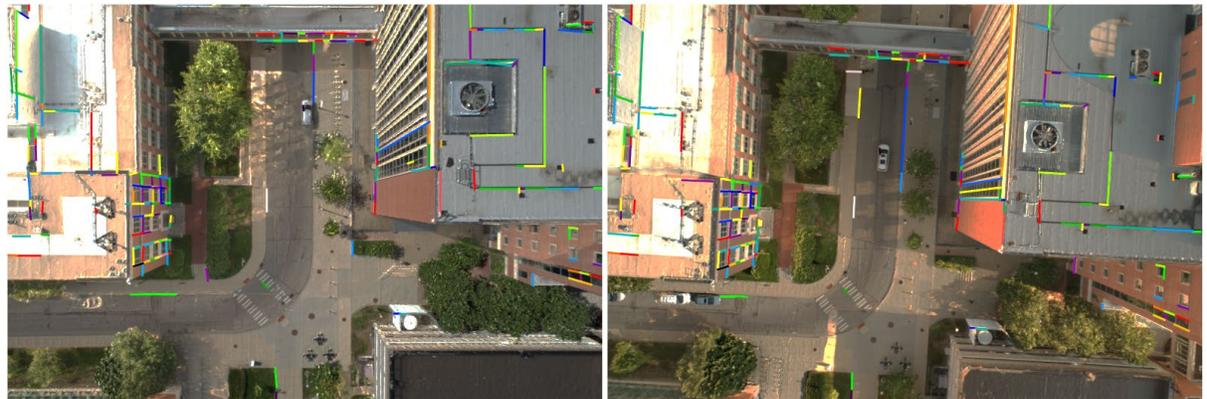

(b) Matched line segments of our albedo pair

Figure 17. SOLD2 (Pautrat et al., 2021) line segments matching results using original images and our albedo decomposition.

### 6.3 *Application: stereo and multi-view dense image matching*

Similarly, more consistent images may yield better dense image matching results. Therefore, our proposed albedo recovery method may augment the dense matching results. Here, by using images before and after our albedo correction method (Figure 18 (a) and (c)), we perform a multi-view stereo dense image matching using the Patch-Match algorithm (Barnes et al., 2009) followed by a multi-view depth map fusion, both are implemented by OpenMVS (Cernea, 2020), a very commonly used open-source software package. The resulting point clouds are shown in Figure 18. To verify that the effectiveness of our method is more than just a brightness balancing in the shadowed region. We have also performed a set of experiments by performing a gamma correction (Smith, 1995) prior to dense matching, results shown in Figure 16(b). It is obvious that our albedo recovered image set yields more details and higher completeness overall.

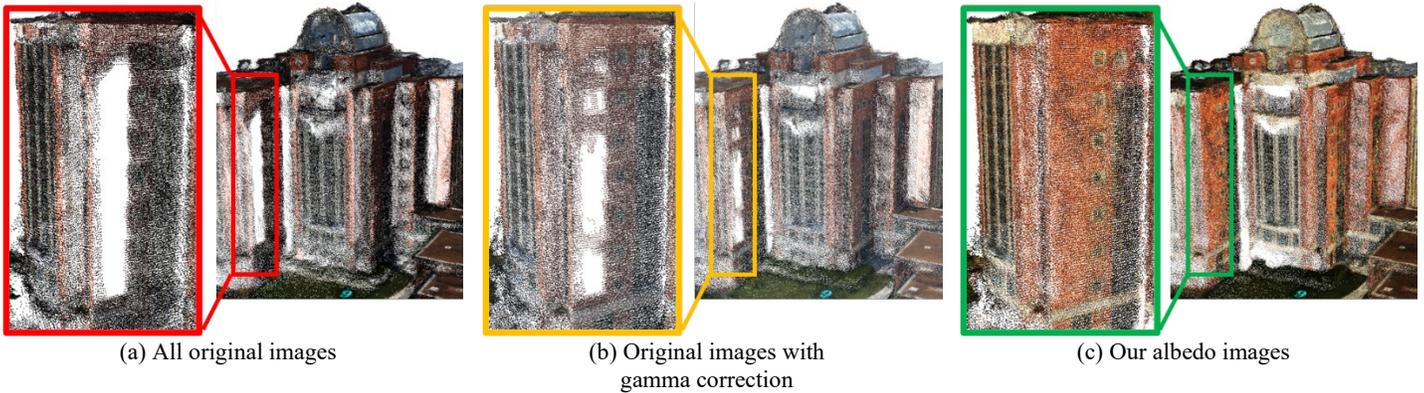

(a) All original images　　　(b) Original images with gamma correction　　　(c) Our albedo images

Figure 18. Comparison of multi-view stereo point clouds generated from all images (290 images) of the region. For gamma correction (b), we scale the colors by a power of 1/2.2. Colored boxes indicate the region where our albedo point cloud presents better completeness.

### 6.4 *Application: change detection*

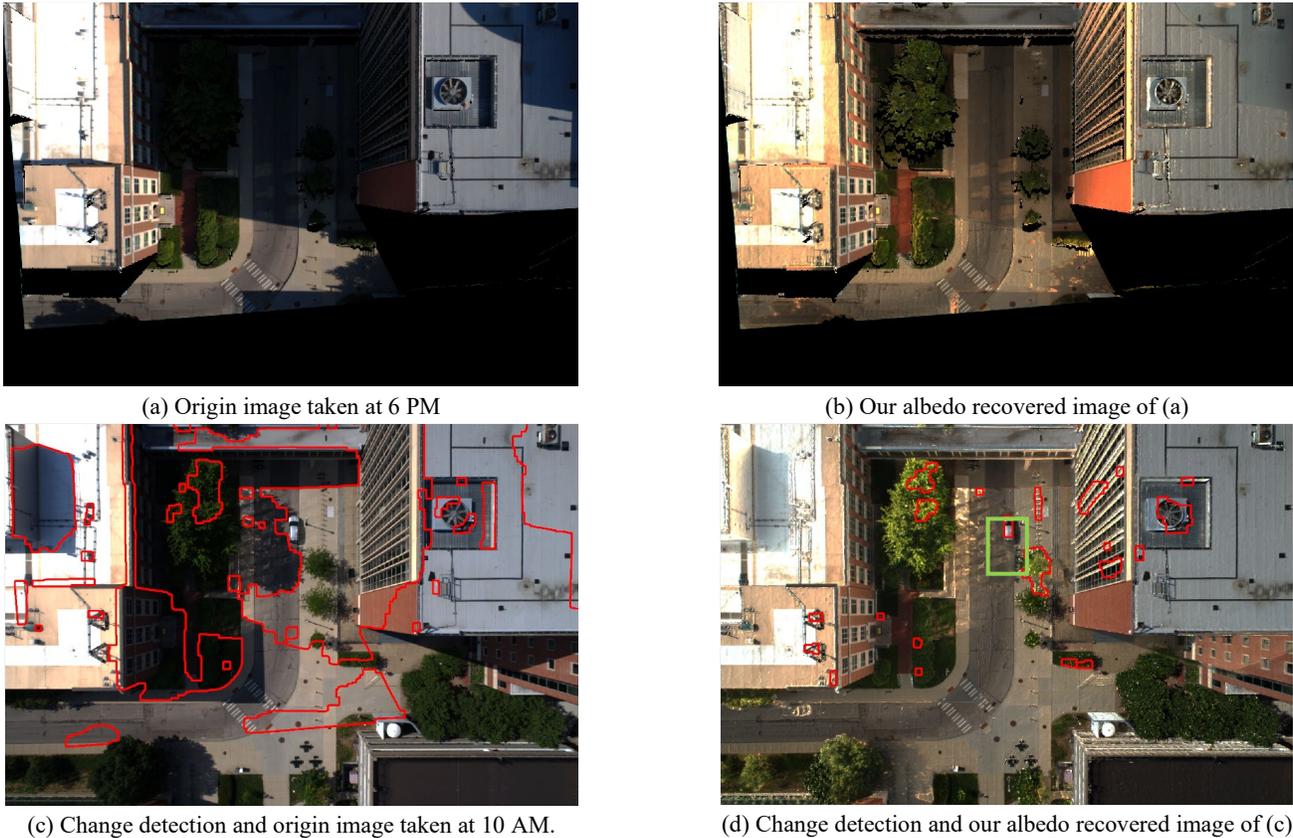

(a) Origin image taken at 6 PM　　　(b) Our albedo recovered image of (a)

(c) Change detection and origin image taken at 10 AM.　　　(d) Change detection and our albedo recovered image of (c)

Figure 19. Change detection across different times of the day. Change mask from Algorithm 2 is visualized as red contours in (c) and (d). All images are geometrically corrected to the same viewpoint. Black regions in (a) and (b) are due to occlusions from view correction. The green box in (d) indicates the changed object between 10 AM and 6 PM.

Shading effects are one of the major challenges in image-based change detection, which adds false positives to change detection algorithms. For example, shadows may be misinterpreted as changes or actual changes may be obscured under the shadowed. Thus, if the shading effects can be reduced, changes could be easily detected by comparing overlapped pixels. Here we use a simple pixel-wise image differencing-based change detection algorithm (Algorithm 2) to demonstrate the capacity of our albedo recovery method. Views of different images are corrected to the same viewpoint using the 3D geometry for pixel-to-pixel alignment. Figure 19 shows that the change detection algorithm on our albedo recovered image can detect reasonable changes such as transient or moving objects in the scene (Figure 17(b)), while the results on the original images show that it is heavily polluted by the unwanted shadows (Figure 17(c)). Our method could potentially serve as a pre-processing step for future research on advanced change detection techniques, aimed at enhancing performance, especially under significant illumination differences.

---

**Algorithm 2 A simple change detection method**

1. Correct the multi-temporal images to the same viewpoint using a 3D model and camera poses.
2. Compute image differencing between the source image and reference image.
3. Convert the difference map into a binary mask by applying a threshold.
4. Apply morphological open & close operations to clean out the binary mask.
5. Output cleaned binary mask.

---

## 7. Conclusion

This paper presents a general albedo recovery approach for photogrammetric images, as an extended work from our earlier work (Song and Qin, 2022). The core of this approach is a data-agnostic outdoor light modeling, by taking metadata from photogrammetry data collection, the approach directly estimates the sunlight direction, based on which a heterogenous skylight model is estimated by utilizing lit-shadow samples in the image observations and the local geometry. We provide a more comprehensive math framework for light modeling, as well as more comprehensive experimental results demonstrating the effectiveness and scalability of our methods. As compared to existing albedo recovery methods, we show that our proposed method significantly outperforms others under the context of photogrammetric collection both in terms of quantitative metrics (PSNR, SSIM, etc.), qualitative visual comparison, as well as the improvement of downstream applications that the proposed method can drive, including model relighting, feature extraction & matching, dense stereo image matching, and change detection. As compared to many existing works, our approach does not require additional data collection logistics and can be easily scalable. In this paper, the proposed models are developed based on the Lambertian surface assumption. This is a choice that balances between the simplicity of the solver for scaling the algorithm to a practical level, and the accuracy of the models. From a theoretical point of view, the lack of modeling specular reflections may cause problems on reflective surfaces, while practically a simpler (and linear) model would suffer less from outliers caused by failures of non-linear optimization. Our experiments show that the proposed approach on full-scale photogrammetric models with its full-resolution albedo recovery is effective at this proof-of-concept (PoC) stage. In our next attempt at research, we aim to address the incorporation of models beyond Lambertian in the albedo recovery problem under this practical context, which includes models for reflective surfaces, skylight models under weathered conditions, and its corresponding robust solvers.

## Acknowledgment

This work is supported by the Office of Naval Research (Award No. N000142012141 and N000142312670) and IARPA (Award No. 140D0423C0075)